\documentclass[11pt]{article}

\usepackage[final]{acl}
\usepackage{placeins}
\usepackage{times}
\usepackage{latexsym}
\usepackage{enumitem}
\usepackage{booktabs}
\usepackage{diagbox}
\usepackage{makecell}
\usepackage{multirow}
\usepackage[T1]{fontenc}

\usepackage[utf8]{inputenc}

\usepackage{microtype}

\usepackage{inconsolata}

\usepackage{graphicx}
\usepackage{amssymb} 
\title{A Systematic Comparison between Extractive Self-Explanations and Human Rationales in Text Classification}

\author{Stephanie Brandl\thanks{\hspace{2mm} Equal contribution.} \\
  Center for Social Data Science \\ 
  University of Copenhagen \\
  \texttt{stephanie.brandl@sodas.ku.dk} \\\And
  Oliver Eberle$^{*}$ \\
  Machine Learning Group \\ 
  Technische Universität Berlin\\
  \texttt{oliver.eberle@tu-berlin.de} \\}

\begin{document}
\maketitle
\begin{abstract}
Instruction-tuned LLMs are able to provide \textit{an} explanation about their output to users by generating self-explanations, without requiring the application of complex interpretability techniques. In this paper, we analyse whether this ability results in a \textit{good} explanation. We evaluate self-explanations in the form of input rationales with respect to their plausibility to humans. We study three text classification tasks: sentiment classification, forced labour detection and claim verification. We include Danish and Italian translations of the sentiment classification task and compare self-explanations to human annotations. For this, we collected human rationale annotations for Climate-Fever, a claim verification dataset.
We furthermore evaluate the faithfulness of human and self-explanation rationales with respect to correct model predictions, and extend the study by incorporating post-hoc attribution-based explanations. We analyse four open-weight LLMs and find that alignment between self-explanations and human rationales highly depends on text length and task complexity. Nevertheless, self-explanations yield faithful subsets of token-level rationales, whereas post-hoc attribution methods tend to emphasize structural and formatting tokens, reflecting fundamentally different explanation strategies.

\end{abstract}

\section{Introduction}

Providing model explanations to increase trustworthiness and transparency is a key goal of interpretability research, with LLMs offering new ways to trace model decision-making. As AI-based analyses increasingly have the power to influence decisions and perceptions, systematically tracing and comparing explanations of these decisions is essential for fostering trust, transparency and regulatory compliance.
Today, LLMs are being used for a wide variety of tasks, ranging from creative writing and homework assistance to offering advice, summarizing reports and translation, while providing self-generated explanations of their outputs, i.e., self-explanations, in the process.\footnote{\href{https://www.washingtonpost.com/technology/2024/08/04/chatgpt-use-real-ai-chatbot-conversations/}{\nolinkurl{www.washingtonpost.com/technology/2024/08/04/chatgpt-use-real-ai-chatbot-conversations}}} This makes it even more important to understand the quality of those self-explanations, how reliable they are and to evaluate their faithfulness to the model and their plausibility to humans.
\begin{figure}
    \centering
    \includegraphics[width=0.44\textwidth]{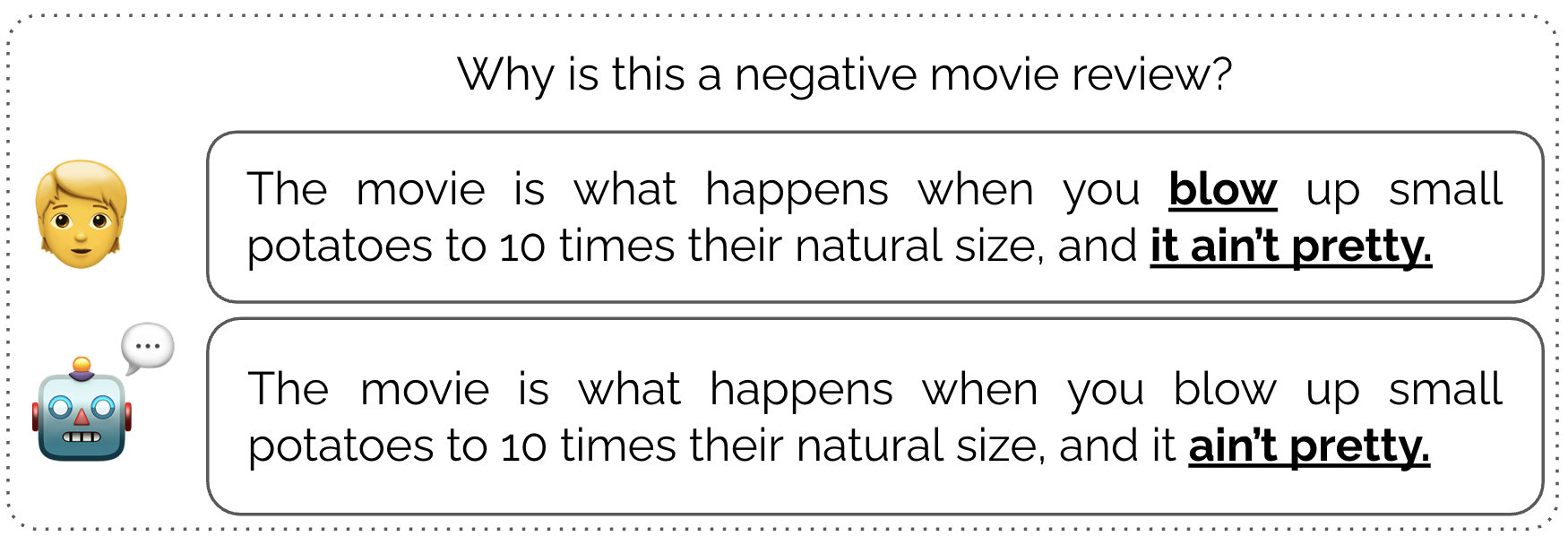}
    \caption{An example from the SST sentiment classification dataset. With rationale annotations by humans and generated by \textit{Llama3}.}
    \label{fig:main}
    \vspace{-15pt}
\end{figure}
In this paper, we evaluate self-explanations from two text classification tasks for which human rationale annotations are available: sentiment classification and forced labour detection, i.e., identifying pre-defined risk indicators of forced labour. We also collect and release new rationale annotations by three personally instructed annotators for a subset of Climate-Fever, a claim verification dataset with claims related to climate change. We instruct 4 different LLMs (Gemma3, Llama3, Mistral and Qwen3) to solve the respective tasks and generate rationales based on the input text in a zero-shot experiment. 
For sentiment classification, we consider two different subsets from two different annotation studies, one also including Italian and Danish translations alongside the English original. 
Following established evaluation methods in the interpretability and Explainable AI  literature \cite{deyoung-etal-2020-eraser, jacovi-goldberg-2020-towards}, we assess the plausibility of model rationales by measuring their agreement with human annotations and evaluate faithfulness via interventions on input tokens to determine their importance for the model’s decision.
Our study takes an initial step towards a better understanding of the reliability and quality of self-explanations, for which we analyzed four language models, three languages, and three distinct text classification domains. Those domains widely vary in text length and task complexity and are more realistic in how they are composed than the datasets in many other explainability studies. While most studies focus on short texts and keyword-based tasks, we also include tasks such as forced labour detection on news articles and claim verification on automatically retrieved text snippets from Wikipedia where neither the task itself nor the rationale annotation is trivial, requiring  domain-specific terminology and non-trivial evidence assessment.
In order to ensure a fair comparison, we consider state-of-the-art gradient-based attribution methods to compute post-hoc explanations which have not been systematically compared with self-explanations so far.
Our findings provide relevant insights for model interpretability and user trust in self-explanations, and we further support reproducibility and future research by openly releasing our code and the collected rationale annotations.\footnote{\href{https://github.com/oeberle/self_explanations_human_rationales}{\nolinkurl{https://github.com/oeberle/self_explanations_human_rationales}}}\footnote{\href{https://huggingface.co/datasets/stephaniebrandl/climate_fever_rationales}{\nolinkurl{https://huggingface.co/datasets/stephaniebrandl/climate_fever_rationales}}}

\textbf{Contributions} In this work, we
\begin{enumerate}[label={(\roman*)}, noitemsep, labelsep=4pt, noitemsep, topsep=0pt, parsep=0pt, leftmargin=16pt]
    \item collect human rationales for a subset of Climate-Fever, a climate claim verification dataset.
    \item conduct a \textit{controlled study comparing human annotations with LLM-generated explanations}.
    \item evaluate \textit{plausibility}, i.e.,the level of agreement between model and human rationales and \textit{faithfulness}, i.e., the relevance of selected rationale tokens for the task (model decision).
    \item study three different text classification tasks: \textit{sentiment classification}, \textit{forced labour detection} and \textit{claim verification}.
    \item  systematically compare human, model, and post-hoc attribution-based rationales, showing self-explanations do yield faithful token-level explanations, whereas post-hoc attributions emphasize structural and formatting tokens, reflecting different explanation strategies.
\end{enumerate}
\vspace{5pt}

\textbf{Extractive vs.~abstractive explanations.}
Extractive explanations such as token-level rationales, have been the standard setting in the interpretability and Explainable AI literature, e.g., in the context of sentence labeling, (closed-book) question answering, or factual recall. In this study, we have chosen to focus on extractive explanations, as the explanation is grounded in the provided input, and thus can be clearly evaluated with respect to human rationale annotations. Relevant evaluation methods for such extractive settings have been established over many years, whereas the evaluation of free-text explanations for faithfulness and plausibility are still being developed \cite{ye2022unreliability, wiegreffe-etal-2022-reframing, kunz-kuhlmann-2024-properties, madsen-etal-2024-self}. We see this controlled analysis as a first step into building reliable self-explanations. Abstractive evaluation of generative AI presents a much broader challenge in evaluating generative AI outputs and explanations, see also \citet{ross2021evaluating, sarti2023inseq}.

\section{Related Work} \label{app:related_work}
Generated self-explanations present both new opportunities and challenges. Prior work in self-explanations for text has focused on new evaluation strategies and model improvements. \citet{ye2022unreliability} evaluate whether including self-explanations can improve model performance on in-context learning while \citet{resck-etal-2024-exploring} incorporate human-annotated rationales during model training to improve plausibility of post-hoc explanations in text classification. Other work focuses on instruction-based self-consistency checks to measure faithfulness in different types of generated explanations \cite{madsen-etal-2024-self, parcalabescu-frank-2024-measuring,zhao-iii-2025-necessary}. 


Another line of work by \citet{wiegreffe-etal-2022-reframing} seeks to improve free text self-explanations with the help of human-written explanations that are included in the instruction. Similarly to \citet{kunz-kuhlmann-2024-properties}, self-explanations are evaluated on a variety of properties by the means of human annotation. They are found to be generally true, grammatical and factual \cite{wiegreffe-etal-2022-reframing} and further selective, to contain illustrative examples and rarely subjective according to \citeauthor{kunz-kuhlmann-2024-properties}. 
\citet{wang-atanasova-2025-self} explore different self-feedback strategies to improve free text explanations and find that extracting rationales from the input to be the best among their strategies.

In our study, we consider human rationales as the ground truth for explaining a decision, against which we compare model self-explanations in order to evaluate how plausible they are.

Recent work by \citet{huang2023largelanguagemodelsexplain} investigates self-explanations by ChatGPT on sentiment classification for SST, comparing faithfulness of self-explanations against different feature attribution methods. They experiment with different settings by swapping the order of classification and explanations within a single instruction prompt, asking the model for top-k rationale tokens or continuous token scores, but find no method that stands out in faithfulness while observing significant disagreement across explainability approaches. 
\citet{randl2025mind} compare extractive self-explanations from three text classification tasks with rather short texts with saliency-based explanations and human rationale annotations. They exclude the more sophisticated gradient-based methods in their work.

Instead, we ground our evaluation in human-annotated rationales to assess plausibility, and use post-hoc LRP attributions as well as GradientxInput specifically designed for Transformer models to measure faithfulness \cite{transformerxai2022, pmlr-v235-achtibat24a}. This systematic approach defines a clear evaluation setting to effectively assess explanation quality by comparing attribution-based explanations, self-explanations, and human rationales alongside a random baseline. The three text classification tasks we include cover a wider range of difficulty in terms of ambiguity, multilinguality and terminology than previous work.

\section{Experimental Setup}

\subsection{Datasets}
We select two text classification datasets for sentiment analysis and forced labour detection, for which human rationale annotations had been collected. We also collect human rationales for a third dataset which contains a claim verification task.
With those three datasets we cover different aspects and levels of difficulty in both classification and rationale annotation. 
SST has been widely used for binary sentiment classification, with rationales available in English, Italian and Danish subsets. Texts are rather short and language models have been shown to solve this task successfully, while the second dataset of longer news articles on forced labour detection is more challenging for both classification and rationale extraction, and is also less likely to have been part of the models' pre-training.\\
We collect new human rationales for a subset of \textit{Climate-Fever}, an English claim verification dataset of 1.535 real-world claims about climate change with 5 evidences each (7.675 in total), automatically retrieved from Wikipedia. Here, claims are often ambiguous and it is sometimes unclear if evidences refer to the claim or a semantically similar topic which makes it challenging to classify both evidences and claims for models and humans.


\textbf{SST/mSST}
We use two different subsets from the \textbf{S}tanford \textbf{S}entiment \textbf{T}reebank (SST2, \citealt{socher-etal-2013-recursive}) for binary sentiment classification on movie reviews. The first subset (SST) contains 263 English samples from the validation and test split from SST2 with an average sentence length of 18 tokens. Human rationale annotations have been published for that subset by \citet{thorn-jakobsen-etal-2023-right} where each sample has been annotated by multiple annotators, 8 on average, who were recruited via Prolific. Annotators were first asked to classify the sample into one of three classes: \textit{positive, neutral} or \textit{negative} where none of the sentences was assigned \textit{neutral} as a gold label. In a second step, annotators should choose the parts of the input that support their label choice. We select the rationale annotations with the correct labels from the first step for further analysis. We averaged the binary rationales across all annotators (with correct label classification) and set a threshold of $0.5$ (after averaging) for the token selection. We additionally analyse the rationale annotations collected by \citet{jorgensen-etal-2022-multilingual} on a subset of 250 samples from the validation set of SST2 (mSST). All samples were translated into Danish and Italian with an average sentence length of 15-17. Rationale annotation was carried out by 2 annotators per language (including English), who were native speakers with linguistic training. In contrast to the annotations collected by \citeauthor{thorn-jakobsen-etal-2023-right}, the correct sentiment (\textit{positive} or \textit{negative}) was provided and the annotators were asked to select parts of the input that supported the gold label. 

\textbf{RaFoLa} The authors of \citet{mendez-guzman-etal-2022-rafola} published a \textbf{Ra}tionale-annotated corpus for \textbf{Fo}rced-\textbf{La}bour detection. This multi-class and multi-label dataset contains 989 English news articles that were labeled and annotated according to 11 risk indicators defined by \citet{ilo2012indicators}. Rationale annotations were carried out by two annotators who selected parts of the input to justify their label decision if they found evidence for any of the 11 indicators. A subset of 100 articles was annotated by both annotators with a label agreement of $0.81$ (micro F1) and a rationale agreement of $0.73$ (intersection-over-union). The remaining articles were only annotated by one of the annotators. Each news article was assigned $1.2$ labels on average while $43\%$ were assigned with at least one label. For our analysis, we selected the 4 most frequent classes with occurrences between 117-256 out of the 989 articles. As we carry out zero-shot experiments on models that have not been fine-tuned on this task, we further convert this task into a binary classification task where we ask for a specific label once at a time. We provide the definition of the respective forced labour indicator as part of the instruction, see Figures \ref{fig:rafola_instructions} \& \ref{fig:rafola_definitions}. Grounded in internationally defined labor standards, the forced labour risk indicators in this dataset serve as a socially relevant case study and valuable benchmark for evaluating how well meaningful and faithful explanations can be extracted from language models in a challenging real-world classification task.

\textbf{Climate-Fever} First published by \citet{Climate-Fever}, this dataset contains claim-evidence pairs with real-world claims retrieved from the web and 5 evidences for each automatically retrieved as text snippets from Wikipedia. The original evidences were labeled by annotators (\textit{support}, \textit{refute} or \textit{not enough info}) and based on a majority vote across all 5 evidences, the overall claim was labeled as either \textit{support}, \textit{refute}, \textit{not enough info} or \textit{dispute}. We manually select 104 claims with 520 evidences, aiming for a balanced label set and claims that are easy to understand. Three personally recruited annotators then annotated rationales for the 520 evidences that either support or refute the claim or are left without label if not relevant. Pairwise inter-annotator agreement across all tokens for the 3-class settings reaches an average of $0.36\pm 0.05$ (Cohen's Kappa).
For further analysis, rationales were averaged with the same strategy as for SST with thresholds of $0.5$ and $-0.5$. 
Figure \ref{fig:polarbear} in the Appendix illustrates the difficulty of annotating this dataset. The claim itself \textit{The polar bear population has been growing} is under-specified, it neither refers to a specific time frame nor a geographic location. On the other hand, among the automatically retrieved evidences, we find specific time frames (\#2 and \#3) but can sometimes only assume that a particular statement refers to the original claim (\#3 mentions \textit{bears} but not \textit{polar bears}). More details on the annotation study can be found in  Appendix\ref{sec:annotation}. 

\subsection{Rationale Extraction}
For our experiments, we evaluate the following 4 instruction fine-tuned LLMs: Gemma3-12B (language-only), Llama3.1-8B, Qwen3-8B (non-thinking mode) and Mistral-7B with more details in Appendix \ref{sec:models}.

We first ask the model to classify the given text into positive/negative for SST and into yes/no depending on evidence for a specific risk indicator for the RaFoLa dataset. If the model returns the correct answer, we ask it to generate rationales based on the relevant context of the input. In case of RaFoLa, we follow the original data collection and only request rationales if the respective risk indicator is present.
For the subsets in Italian and Danish, we manually translated the prompts to the respective language with the help of native speakers.

For Climate-Fever, we follow the original annotation approach for both model experiments and human annotations where we first ask for rationales on the 5 provided statements, i.e., evidences with respect to the claim and based on those, ask for an overall claim label.

\textbf{Experimental Details} The experiments are based on the \texttt{transformers} library. We set the repetition penalty to 1.0 and adjust the maximum length of generated text with respect to the task and expected output. We ensure reproducibility of our results by consistently using the same set of 3 seeds across our experiments and will release our code upon publication, including all parameters and the libraries used. Instructions with class definitions are presented in the Appendix in Figures \ref{fig:prompts_sst} - \ref{fig:rafola_definitions}.

\begin{table}[h]
\centering
 \footnotesize
\addtolength{\tabcolsep}{1.5em}
\setlength{\tabcolsep}{1.8pt} 
\begin{tabular}{l|c|c|c|c}
\toprule
\diagbox{Data}{Acc.}& Gemma3 & Llama3 & Qwen3 & Mistral \\
\midrule
SST  & 0.98 & 0.98 & 0.98 & 0.99 \\
\midrule
mSST (EN) & \textbf{1.00} & 0.98 & 0.99 & 0.98\\
mSST (DA) & 0.94 & 0.84 & \textbf{0.96} & \textbf{0.96}\\
mSST (IT) & \textbf{1.00} & 0.95 & \textbf{1.00} & 0.97\\
\midrule
RaFoLA  \#1 & 0.25 & 0.47 & 0.38 & \textbf{0.57}\\
RaFoLA  \#2 & 0.37 & \textbf{0.60} & 0.47 & 0.58\\
RaFoLA  \#5 & \textbf{0.79} & 0.73 & 0.74 & 0.60\\
RaFoLA  \#8 & 0.65 & \textbf{0.76} & 0.67 & 0.73\\
\midrule
\midrule
Claim & \textbf{0.45} $\pm$ .04 & $0.33 \pm .04$ & $0.38 \pm .02$ & $0.24 \pm .01$\\ 
Evidence & \textbf{0.54} $\pm .02$ & $0.40 \pm .03$ & $0.46 \pm .00$ & $0.45 \pm .02$\\
\bottomrule
\end{tabular}
\caption{Model accuracies (macro F1) for SST, multilingual SST, RaFoLa and Climate-Fever (claim verification task and evidence classification), with highest scores shown in bold. Scores are averaged across 3 seeds, standard deviation for the first three datasets is $\leq 0.01$.}
\label{tab:scores}
\vspace{-5mm}
\end{table}


\section{Main Results}
In the following, we will present results on respective task performances and plausibility, i.e., pair-wise agreement between human annotations and generated rationales.


\subsection{Task performance}
Table \ref{tab:scores} shows task accuracies (macro-F1) for SST, mSST, RaFoLa and Climate-Fever. Macro-F1 scores for SST and mSST are generally high across models ($0.84-1.0$). We can assume that most models nowadays have seen the original English version of SST during training and are thus more familiar with this type of data. From the set of models we consider for this study, only \textit{Mistral} is considered an English-only model. \textit{Qwen3} was pre-trained on Italian and Danish, \textit{Llama3} was pre-trained on Italian and for \textit{Gemma3} we only know that it supports 140+ languages but the technical report does not reveal which languages are included \cite{gemma3technicalreport}. Despite this difference in language exposure, our results show that all models are able to solve the sentiment classification task in Danish and Italian with accuracies comparable to English.

RaFoLa shows more variation across articles and overall lower performance ($0.25-0.79$) than SST/mSST. Performance is lower for Articles \#1/\#2 than for Articles \#5/\#8 (best performances $0.57/0.60$ vs.~ $0.79/0.76$). We also observe that, although released more recently, \textit{Gemma3} and \textit{Qwen3}, on average, perform worse than \textit{Mistral} and \textit{Llama3} in this task, $0.52/0.57$ vs.~$0.62/0.64$.

We present claim verification performance and evidence classification for Climate-Fever in the lower part of Table \ref{tab:scores}, both as macro-F1 scores. \textit{Gemma} performs best in both cases, in the 4-class claim verification task it reaches $0.45$ and $0.54$ in the 3-class evidence classification task.


\begin{figure*}
    \includegraphics[width=0.33\textwidth]{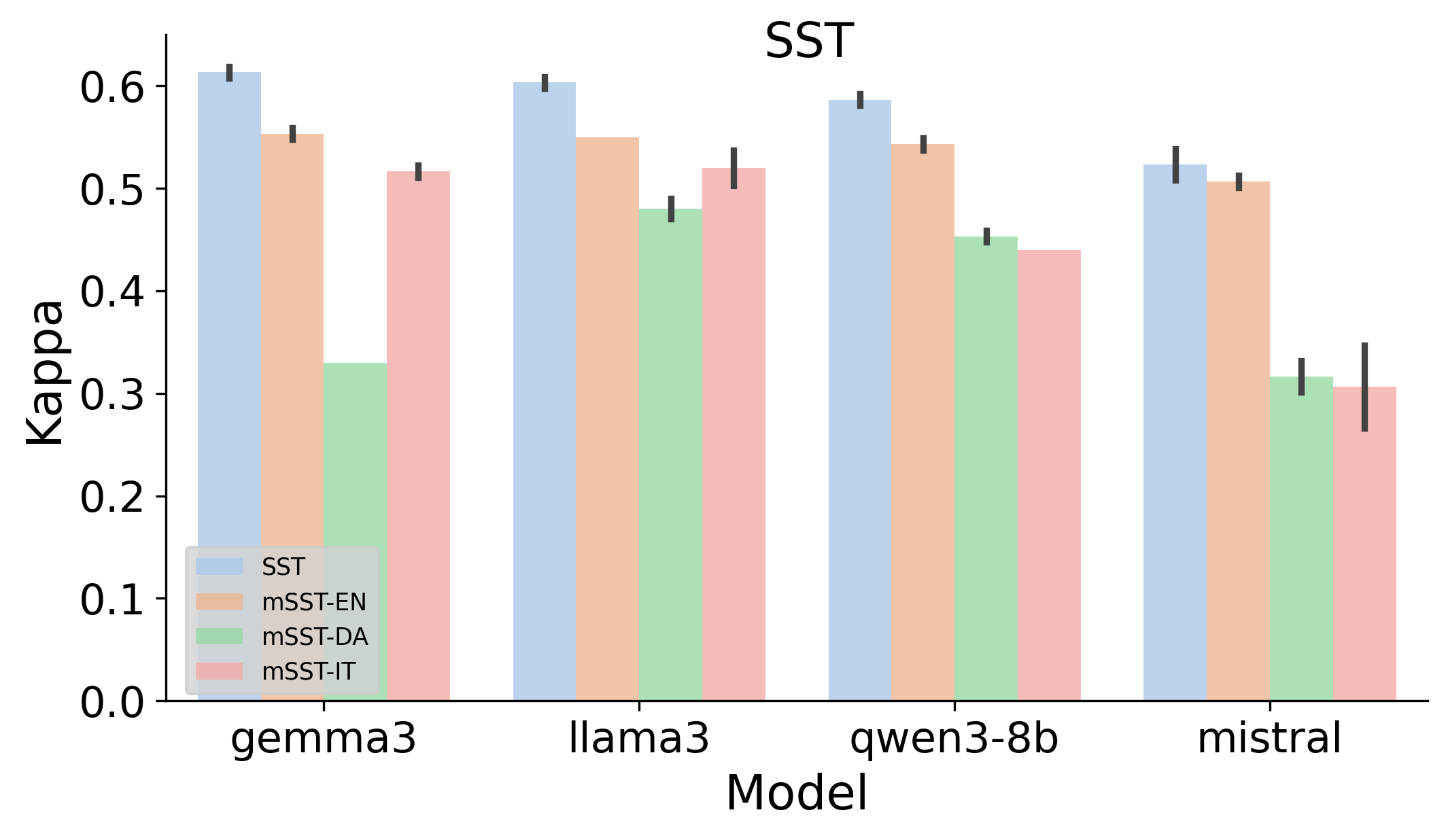}
    \includegraphics[width=0.33\textwidth]{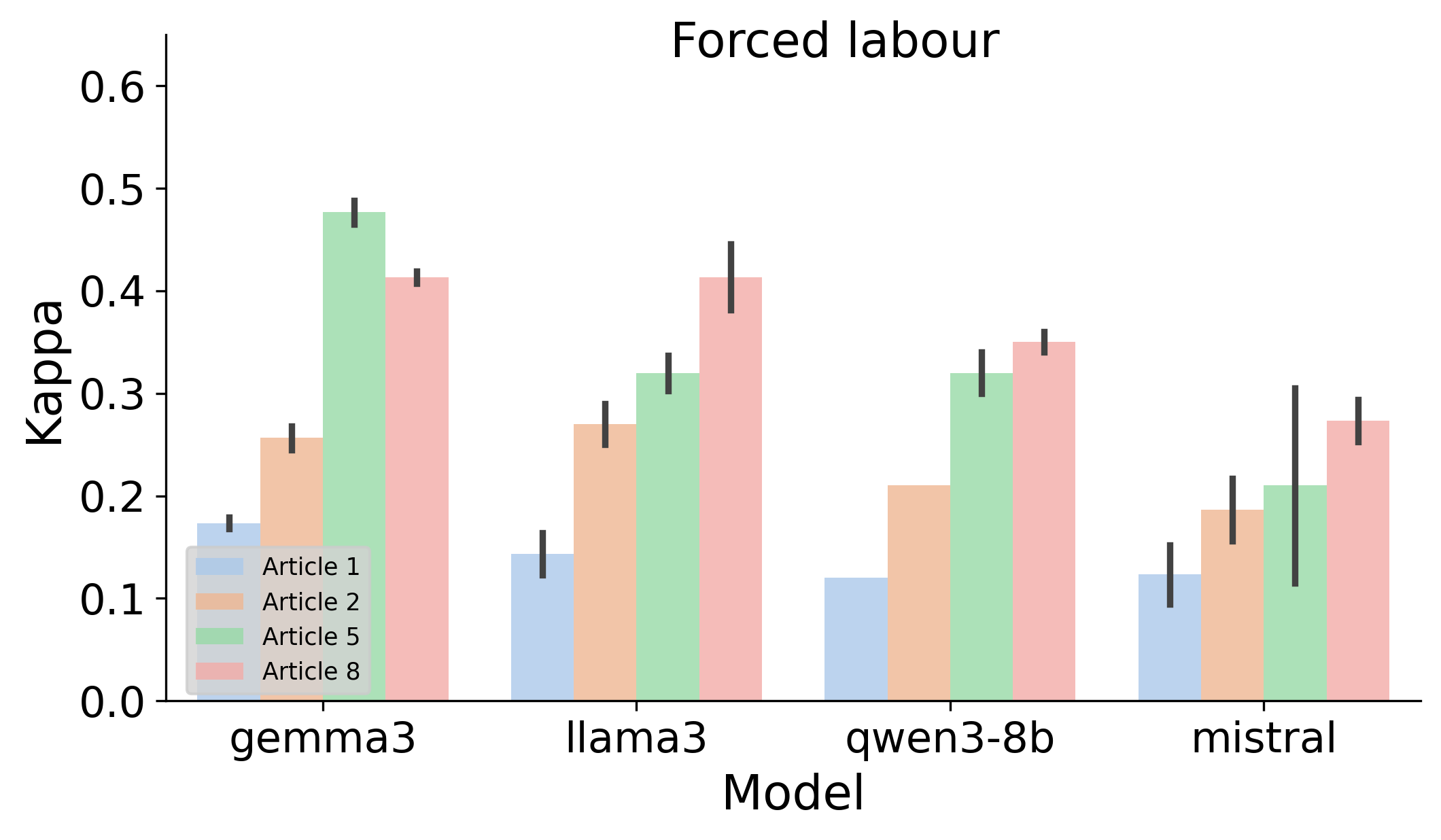}
    \includegraphics[width=0.33\textwidth]{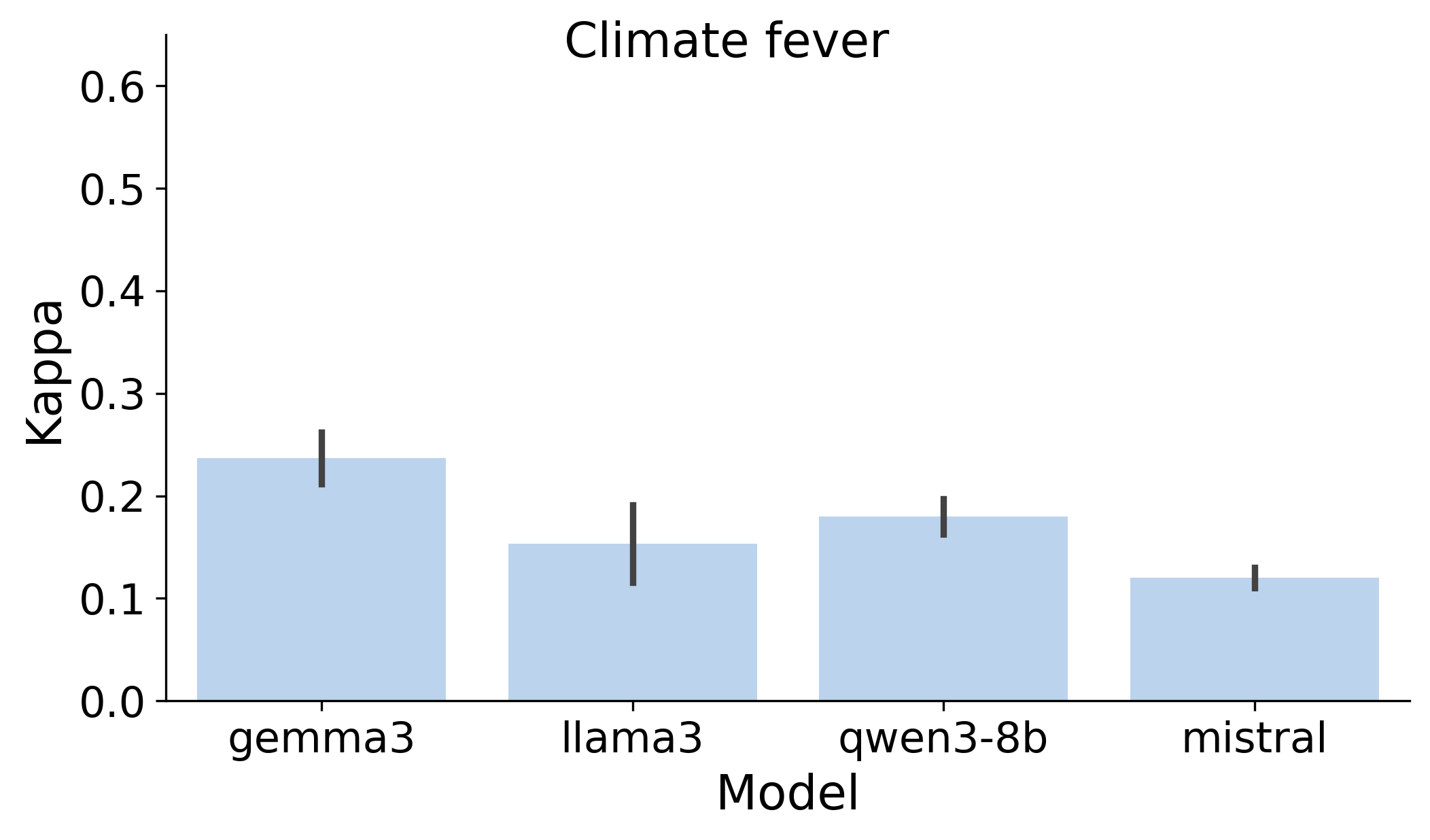}
    \caption{Human-model agreement as Cohen's Kappa scores for all datasets. Scores were computed for correctly classified samples and then averaged across datasets, standard deviation across seeds is shown as error bars.}
    \label{fig:plausibility}
    \vspace{-15pt}
\end{figure*}

\subsection{Plausibility} \label{sec:plausbility}

Following the interpretability literature, we assess \textit{plausibility} of rationales to humans by considering human annotations as the ground truth and compute their agreement to generated rationales \cite{deyoung-etal-2020-eraser}. For this, we calculate sample-wise Cohen's Kappa scores between the binary scores and average across samples for different models. Results (averaged across 3 seeds) are shown in Figure \ref{fig:plausibility}.

\textbf{Metric} Cohen's Kappa \cite{cohen1960coefficient} is a well-established method to measure inter-annotator agreement (IAA) between two annotators, in our case the averaged human annotations and the model rationales. We choose Kappa over F1 scores, which is also often used to evaluate IAA but comes with two obstacles. It is (i) driven by the imbalance of classes (here selected and not selected tokens) leading to a higher offset for annotations with a ratio of selected tokens closer to 0.5 and it (ii) does not consider randomness as a confounding factor. Cohen's Kappa scores account for both issues, leading to overall lower but more robust scores than F1.

\textbf{SST} Results for SST are shown in the left part of Figure \ref{fig:plausibility}. For both English subsets, we mostly see a moderate level of agreement ($0.4 - 0.6$) for the comparison between human annotation and self-explanations (generated by LLMs) except for \textit{Mistral} (DA: $0.32$, IT: $0.31$) and \textit{Gemma} (DA: $0.33$).\footnote{We follow \citet{landis1977measurement} to classify levels of agreement.} 


\textbf{RaFoLa} Results for RaFoLa, are shown in the middle part of Figure \ref{fig:plausibility}. We here see overall lower levels of agreement but also a high variance by a magnitude of up to 4 between different articles. Plausibility scores reach from only lower levels of agreement for article \#1 ($0.12-0.17$), to a fair level of agreement for article \#2 ($0.19-0.27$) and moderate agreements for articles \#5 ($0.21-0.48$) and \#8 ($0.27-0.41$). Similar to SST, we see highest agreements for \textit{Llama3} and \textit{Gemma3} which is surprising given the low performance for \textit{Gemma3}. 


\textbf{Climate-Fever} The right panel of Figure \ref{fig:plausibility} shows Kappa scores for Climate-Fever. Here, we concatenate correctly classified evidence statements per claim and compute agreement scores with human annotations for the respective evidence statements. In contrast to the other datasets, rationales are annotated and generated as either supportive (1), contradictive (-1) or not relevant (0). Only \textit{Gemma3} reaches a fair level of agreement ($0.24$) while the other models reach only slight levels of agreement ($0.12 - 0.18$).

\subsection{Token Statistics} 
We extract the top-8 tokens from all datasets as well as from the rationale annotations. For SST/mSST (Table \ref{tab:top8-SST}) and Climate-Fever (Table \ref{tab:top8-climate}), we do not find any meaningful differences between humans and models, neither across models nor languages. For Climate-Fever we find mostly \textit{nouns} among the selected tokens whereas for SST/mSST we find an even distribution among nouns, adjectives and adverbs. For RaFoLa (Table \ref{tab:top8-Rafola}), we see that articles \#1 (Abuse of vulnerability) \& \#2 (Abusive working and living conditions) mostly contain descriptive nouns covering the general topic of the dataset such as \textit{workers}, \textit{work}, \textit{labour} etc. Whereas selected tokens for articles \#5 (Excessive overtime) \& \#8 (Physical and sexual violence) deviate more from the corpus' most frequent tokens (first row) with keywords such as \textit{hours}, \textit{day} and \textit{sexual}, \textit{women}, \textit{violence}, respectively for the two articles. Those keywords are easier to identify which supports the higher performance for \#5/\#8 observed in Table \ref{tab:scores}.

\paragraph{Summary} We find that (i) task performance and human-model agreement vary a lot across models and articles for RaFoLa with \textit{Llama3} and \textit{Mistral} performing better on average than \textit{Gemma3} and \textit{Qwen3} where the former can be explained by more indicative keywords for some articles than for others, (ii) \textit{Gemma3} outperforms the other models on claim verification and evidence classification for Climate-Fever and (iii) \textit{Gemma3} shows highest agreement with human annotations on all datasets followed by \textit{Llama3}.

\section{Analyses}
In this section, we compare human and model rationales with post hoc interpretability methods to gain a deeper understanding of (i) the extent to which human and model rationales provide faithful token identification, i.e., whether models are sensitive to changes in their predictions when these rationales are masked, (ii) the degree to which human and model rationales align with post hoc attributions, and (iii) how these approaches differ in their strategies for information extraction and token selection. We exclude Climate-Fever from this analysis as there is no clear protocol for applying gradient-based methods on a collection of independent statements.

\begin{figure}[ht!]
    \includegraphics[width=0.98\linewidth, trim={0.6cm 1.1cm 0cm 0.6cm},
  clip]{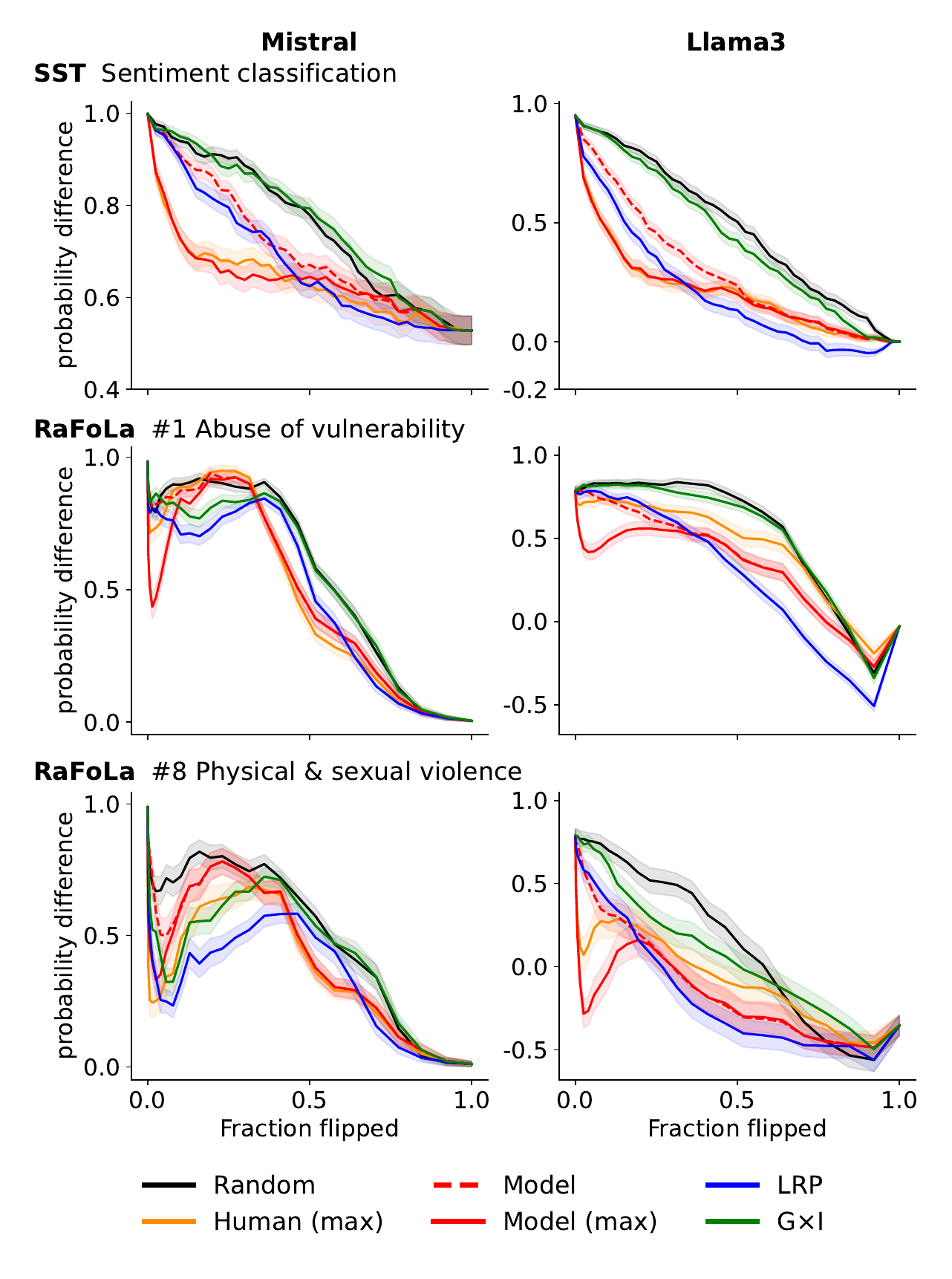}
    \caption{Faithfulness evaluation for SST and RaFoLa (articles \#1 and \#8). Model probability difference after masking tokens extracted from human rationales, model self-explanation rationales and post-hoc attributions (LRP, GxI) for Mistral and Llama3 with full results in Figure \ref{fig:faithfulness_app} (Appendix). Shaded bands indicate standard errors across samples. Faster drop in probability for early fractions indicates more faithful identification of task-relevant rationales. Human/Model (max) refers to rationales selected via greedy maximization of next-token probability difference.}
    \label{fig:faithfulness_main} 
    \vspace{-15pt}
\end{figure}

\subsection{Faithfulness} \label{ref:faithfulness}
Besides plausibility, faithfulness is the most commonly used criterion for evaluating model explanations. It is assessed by removing the most relevant features and measuring the resulting change in the model prediction, which can be viewed as an interventional probe of the model’s decision process. Faithful rationales are characterized by a strong decrease in the prediction score when the associated tokens are iteratively masked. Accordingly, we evaluate faithfulness by measuring the \textit{probability difference} between the correctly predicted answer token and the alternative answer token after masking tokens identified by human rationales, model self-explanations, and post-hoc attributions.

\textbf{Experimental Setup}  To extract input attribution scores, we use two widely adopted gradient-based attribution methods that enable efficient computation of feature attributions in LLMs: Gradient×Input \cite{baehrens10a, GI} and layer-wise relevance propagation (LRP) \cite{transformerxai2022, pmlr-v235-achtibat24a}. Both methods have been successfully applied in recent interpretability work on information retrieval \cite{NEURIPS2024_d6d0e41e} and causal circuit discovery \cite{syed-etal-2024-attribution, jafari2025relp}, and scale effectively to larger models and longer input sequences, allowing us to study more complex datasets like RaFoLa. Details on implementation of attribution techniques are given in Section \ref{app:faithful_rationales} in the Appendix.

We follow a perturbation-based masking approach and rank binary human and model rationales using a k-greedy importance ordering algorithm that prioritizes tokens according to their maximal impact on the probability difference in descending order.
Iteratively selecting the top-k tokens that produce the largest drop in probability difference and flipping them first yields a ranking that captures the cumulative contribution of tokens to the model prediction, which we denote as \textit{Human/Model (max)}. We set $k=1$ for the shorter SST sentences, and $k=3$ for all other datasets. Our faithfulness results on SST and RaFoLa subsets for \textit{Mistral} and \textit{Llama3} are summarized in Figure \ref{fig:faithfulness_main}, with results for all models provided in Appendix~Figure~\ref{fig:faithfulness_app}.


\textbf{Ranked self-explanations offer faithful model interventions.} 
Steepest drop in probability difference occurs for k-greedy ranked self-explanations (Model (max)), especially for the first 5-10\% of tokens perturbed. Flipping of model self-explanations in random order (dashed red line) results in less pronounced impact on probability difference without a pronounced dip in probability difference.

\textbf{Self-explanations can identify more faithful rationales than human annotators.} We find that model rationales extracted from self-explanations are overall at least as faithful as human rationales, in particular, considering the first 0-10\% of token fractions flipped. Specifically, we observe a transient drop followed by a rebound in probability difference for long-text classification settings in RaFoLa compared to SST sentiment classification. 
This \textit{faithfulness dip} can also occur for human rationales, but is less pronounced. As RaFoLa requires weighting potentially contradictory evidence in news articles, removing highly influential phrases can alter the meaning of other evidence. This suggests more complex text interactions than in short sentiment classification, potentially reflecting higher-order semantic and epistemic processes that drive the observed rebound.

\textbf{Post-hoc explanations differ from both humans and model rationals.} 
Depending both on task and model, we see good to moderate agreement with self-explanations, in particular, for SST and RaFoLa \#8, and especially for the first 10\% of flipped tokens. However, we find clear differences in faithfulness curves, as models react differently to interventions on model versus post-hoc (LRP) rationales for RaFoLa. LRP-based rationales provide consistently more faithful token identification than GradientxInput, while overall, post-hoc attributions being less faithful in detecting the first few most relevant tokens than model self-explanation rationales. While post-hoc LRP rationales can, depending on the model, result in fast drops in probability difference, its ordering remains less faithful than both model and human rationales, hinting at differences in how text importance is assigned.


These differences may reflect the \textbf{distinct interpretability approaches}, with model rationales relying on coherent evidence communicated through natural language explanations while post-hoc methods assign relevance to the entire input. The most attributed tokens are often not task-specific content (e.g., RaFoLA news articles) but structural tokens from the system prompt, such as "<|begin\_of\_text|>" in Llama3, "<s>" in Mistral, or "<bos>" in Gemma3. As these tokens are required for the model’s internal processing and sequence delimitation, attribution methods mark them as highly important despite their limited semantic role. Consequently, post-hoc rationales tend to emphasize structural or formatting tokens rather than evidence, highlighting a key difference between evidence-focused explanations and natural-language model rationales. 

\subsection{Analyzing Rationale Strategies in RaFoLa}\label{ref:rat_strategies}

We examine the top 5\% of faithful tokens obtained via intervention-based analysis on the RaFoLA dataset, where faithfulness curves differ most across human, model, and post-hoc rationales. Through an initial LLM-assisted exploratory analysis of top flipped tokens using GPT-5 (see \ref{app:llm_analysis} in the Appendix for details), we identified recurring themes and preliminary distinctions across these groups. We validate those findings with a statistical analyses and standard NLP pipelines. This reveals a variety of strategies that reflect systematic differences in focus and function across rationales.

Overall, we find that \textbf{human rationales contain more tokens that convey narrative content}, emphasizing lived experiences and the broader significance of exploitation through story-like language. This is reflected in higher lexical diversity, with type–token ratios ranging from 29\% to 44\%, and high proportions of stop words, typically 34–39\% across models (see Table~\ref{tab:faithful_rationales} in the Appendix). Typical rationales include descriptive and narrative words like \textit{vulnerable, victims, poverty, children, working, factory, men, women} and phrases like \textit{One year I was pregnant [...]}, \textit{girl describes how her boss [...]}, or \textit{She told me}, reflecting a focus on human contexts and experiences. Tables \ref{tab:rafola_samples_art1} and \ref{tab:rafola_samples_art8} in the Appendix show additional samples.

\textbf{Model-generated rationales provide more factual and analytical detail}, using denser, technical phrasing to document mechanisms and procedures, as well as quantitative evidence. Their lexical diversity is similar or slightly higher than humans (type–token ratios 28–47\%), while stop word fractions remain comparable (35–38\%), indicating natural-language style usage. Example (sub)tokens include \textit{trafficking, labour, bail, conditions, hundreds, \$, Johannesburg, Xinjiang} and phrases like \textit{The Global Slavery Index} or \textit{million labourers [...] is the
biggest recruitment programme anywhere
in the world according to the International
Labor Organization ILO},  highlighting their tendency to record events, locations, and procedural aspects in a factual manner.

\textbf{Post-hoc rationales emphasize more isolated evidence tokens and structural elements}, including publisher artifacts such as source labels, dates, locations, and editorial markers. These rationales contain lower stop word fractions (12–25\%), lower lexical diversity (type–token ratios 24–37\%), and higher proportions of named entities—for example, GPE up to 1.69\%, ORG up to 2.74\%, and PERSON up to 2.23\% compared to human and model rationales. Typical tokens include \textit{Reuters, CNN, "By", Swiss, Lisa Krist, Ghana, "https:/"}, reflecting their focus on evidence extraction and sources.

Differences relative to human rationales are quantified by the $\Delta$ values reported in Table~\ref{tab:faithful_rationales}, which show both model vs.~human and post-hoc vs.~human difference patterns. Positive or negative $\Delta$ values indicate increases or decreases, respectively, in token-type ratios, stop word fractions, formatting tokens, and named entity mentions. For instance, post-hoc rationales typically exhibit 12–24\% lower percentage points of stop word fractions and  higher fractions of named entities (up to 1.51\% for ORG and 1.66\% for PERSON) compared to human rationales, reflecting their evidence-focused role in extracting and isolating token patterns. In contrast, model rationales differ less from human ones, with only modest $\Delta$ values across most linguistic and entity features, emphasizing their adherence to natural-language.

\subsection{Corpus Statistics}
Table \ref{tab:corpus_stats} summarizes key corpus statistics to allow analyzing potential implications regarding the rationale agreement scores.

\textbf{Document and sentence length.} RaFoLa contains substantially longer documents (945 tokens on average) than Climate-Fever (200) and SST (21), suggesting that increased length and contextual complexity may contribute to annotation difficulty. In contrast, the short length of SST likely constrains contextual variation, which may partially explain its higher agreement scores (0.4–0.6). 

\textbf{Lexical diversity and ambiguity.} By computing token-type rations (TTR), we find that SST exhibits the highest lexical diversity (37\% TTR), whereas RaFoLa and Climate-Fever show much lower diversity (3–5\%), while lexical ambiguity remains similar across datasets (54\%). This suggests that neither lexical ambiguity nor diversity alone explains the notably low agreement observed in Climate-Fever (0.12–0.24).

\textbf{Syntactic complexity}.  Climate-Fever has lower mean dependency depth (MDD = 1.87) than SST (2.75) and RaFoLa (2.92), indicating simpler sentence structures, which combined with moderate document length,  suggests that syntactic complexity is not the primary source of disagreement and instead points toward task-specific effects. 

\textbf{Entities and POS.} Dataset-specific distributions of named entities and POS tags further differentiate the corpora: Climate-Fever contains more quantitative expressions (e.g., cardinals and dates), SST includes relatively few named entities (primarily persons), and RaFoLa shows higher densities of geopolitical, organizational, and demographic entities. These patterns suggest that entity distributions may influence when post-hoc explanations align with or diverge from self-explanations, consistent with our observations in Section \ref{ref:rat_strategies}. 

In summary, our analysis indicates that low agreement scores on Climate-Fever  is not a result of surface-level textual properties and rather relates to its intrinsic task complexity, with SST and RaFoLa exhibiting patterns consistent with their structural and lexical profiles.
\section{Discussion}
In this paper, we evaluate self-explanations, i.e., explanations generated by instruction-tuned LLMs, based on their plausibility to humans and their faithfulness to models. We instruct 4 open-weight LLMs: \textit{Gemma3}, \textit{Llama3}, \textit{Qwen3} and \textit{Mistral} for three text classification tasks in English but also in Italian and Danish. 
We analyse the sentiment classification (SST/mSST) and forced labour classification (RaFoLa) tasks for which human annotations are available. We collect new human rationale annotations for the third dataset, Climate-Fever \cite{Climate-Fever}, a claim verification dataset with claims related to climate change. We further include post-hoc attribution-based explanation methods in order to evaluate faithfulness of rationales to the correct model predictions.

For the \textbf{plausibility} analysis, we compute agreement scores between human and model rationales. Overall, we find highest plausibility scores for the English subsets of SST. For RaFoLa, we find higher variance across the different articles which aligns with the task performance. Agreement scores for Climate-Fever, the only dataset with 3-class rationale annotations, reach overall lower agreement scores than the other datasets. 

There are several potential factors that influence those plausibility scores and their differences. The three datasets considered here were originally composed with varying incentives, see also \cite{eberle-etal-2023-rather} for a related discussion.  The movie reviews in SST/mSST were written with the clear purpose of expressing an emotion and justify a given movie rating. Sentences are quite short and usually contain clear keywords that are easy to identify by both humans and models. The RaFoLa dataset consists of news articles whose main purpose is to report on a given event. Those articles are much longer and information about the potential risk indicators might be implicit and subtle as this was not necessarily the main purpose while writing them. As we see in the analysis of the most frequent words, articles \#5 and \#8 contain much clearer and more distinct keywords than articles \#1 and \#2 which makes both classification and rationale annotation more predictable and consistent.
Climate-Fever, a claim verification benchmark pairing real-world web claims with five semantically relevant Wikipedia statements, involves longer evidence contexts and greater ambiguity, making classification and rationale selection more demanding overall. Incorporating such diverse text sources and types of real-world datasets therefore remains essential for obtaining broader and more realistic evaluation benchmarks.

Our analyses further revealed \textbf{distinct faithfulness patterns} for human, self-explanations and post-hoc rationales: human and, in particular, self-explanations are consistently able to intervene strongly on the model's ability to predict correct answer tokens, especially when considering early fractions of text tokens. In comparison, LRP-based post-hoc rationales also reduce the probability difference, but the drop is generally less pronounced. These differences reflect distinct approaches to interpretability: self-explanations communicate coherent evidence through natural language, directly summarizing the model’s decision process, whereas post-hoc methods assign relevance across the entire input, including system and task prompts, often highlighting structural or formatting cues rather than semantically critical phrases. This makes them less effective at disrupting coherent units of meaning but can provide a complementary, more detailed view of the model’s low-level processing compared to self-explanations.
To bridge this gap,  \textit{interpretability agents} \cite{han2025sageagenticexplainerframework, lermen2025deceptiveautomatedinterpretabilitylanguage} that are designed to utilize  model activation or attribution patterns to provide natural language explanations, which may provide more grounded and faithful insights into the mechanisms underlying model predictions.

Good explanations should faithfully reflect the model’s learned strategy, even if not fully plausible or unintuitive to humans \cite{agarwal2024faithfulnessvsplausibility}. Further, self-explanations can suffer from counterfactuality, producing untruthful rationales for correct predictions \cite{hallucinationsurvey2023},  highlighting the need for rigorous evaluation. We therefore concentrated on extractive rationales grounded in the input text, as this setting enables controlled measurements of plausibility and faithfulness against human annotations. Our results constitute an initial step toward characterizing the reliability of self-explanations in realistic text classification. Future research should examine how these findings transfer to abstractive free-text explanations and to process-oriented analyses of how models integrate and resolve complex, potentially contradictory evidence to arrive at a final decision.

\section*{Limitations}

We acknowledge that annotations may be affected by annotator bias, varying guidelines, and differing expertise, impacting the consistency of rationales. Also the number of annotators and the level of details in the instructions varied across the annotation studies we have considered for this paper. Furthermore, for the forced labour detection, annotations by legal scholars might differ from the ones provided and would also be interesting to compare with model rationales. 

We focus our study on rationales based on the input while free text explanations might provide more useful information and pose the more realistic scenario.

While agreement between human and model rationales may be desired, it has been shown in previous work, that humans do not necessarily prefer human-written explanations in comparison to the ones generated by LLMs in the case of free text explanations \cite{wiegreffe-etal-2022-reframing}.

The high zero-shot performance, especially with SST, may be an effect of data contamination, which is likely part of the training data. We can further not exclude the possibility that rationales or task explanations have been included in the training corpus. 



\bibliography{custom}

\appendix

\section{Climate-Fever Annotation Study}\label{sec:annotation}
The Climate-Fever dataset was first collected and published by \cite{Climate-Fever}. The dataset consists of 1535 real-world English claims about climate change for which 7675 related evidence statements were retrieved from Wikipedia, 5 for each claim. Each evidence statement was labeled by up to 5 annotators into either \textsc{support}, \textsc{not enough info} or \textsc{refute}. Micro (per claim-evidence pairs) and macro labels (per claim) were aggregated afterwards. Labels for claim-evidence pairs were decided on a majority vote (or \textsc{not enough info} on a tie). Macro labels were labeled \textsc{not enough info} by default unless there is \textsc{supporting} or \textsc{refuting} evidence and in case of both, the claim was labeled as \textsc{dispute}.\\
For our study, we are interested in token-level rationales which are not available from the initial publication of Climate-Fever. Therefore, we manually selected a subset of 102 claims (510 claim-evidence pairs) based on clarity of the claim formulation and balanced claim labels.
\subsection{Annotator selection}
We used Prodigy as the annotation interface and initially planned to collect data via Prolific, a crowd-sourcing platform. It turned out more difficult than expected for annotators to open the correct link for Prodigy, i.e., they would need to add their Prolific-id to the link and to understand all parts of the annotation study. After several attempts, we decided to instead personally instruct 2 student annotators and one of the authors for the entire study. 
\subsection{Annotation tasks}
\begin{figure*}
    \centering
    \includegraphics[width=\linewidth]{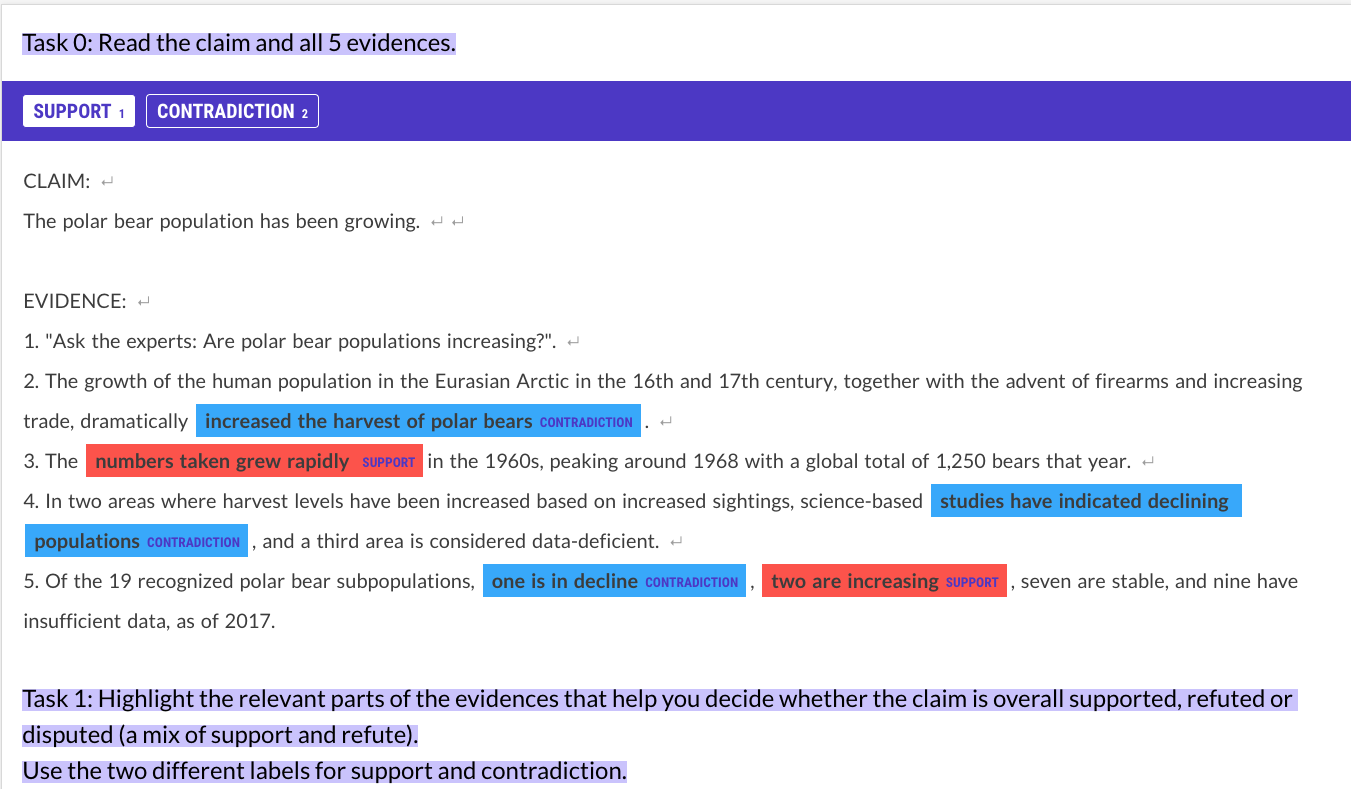}
    \includegraphics[width=\linewidth]{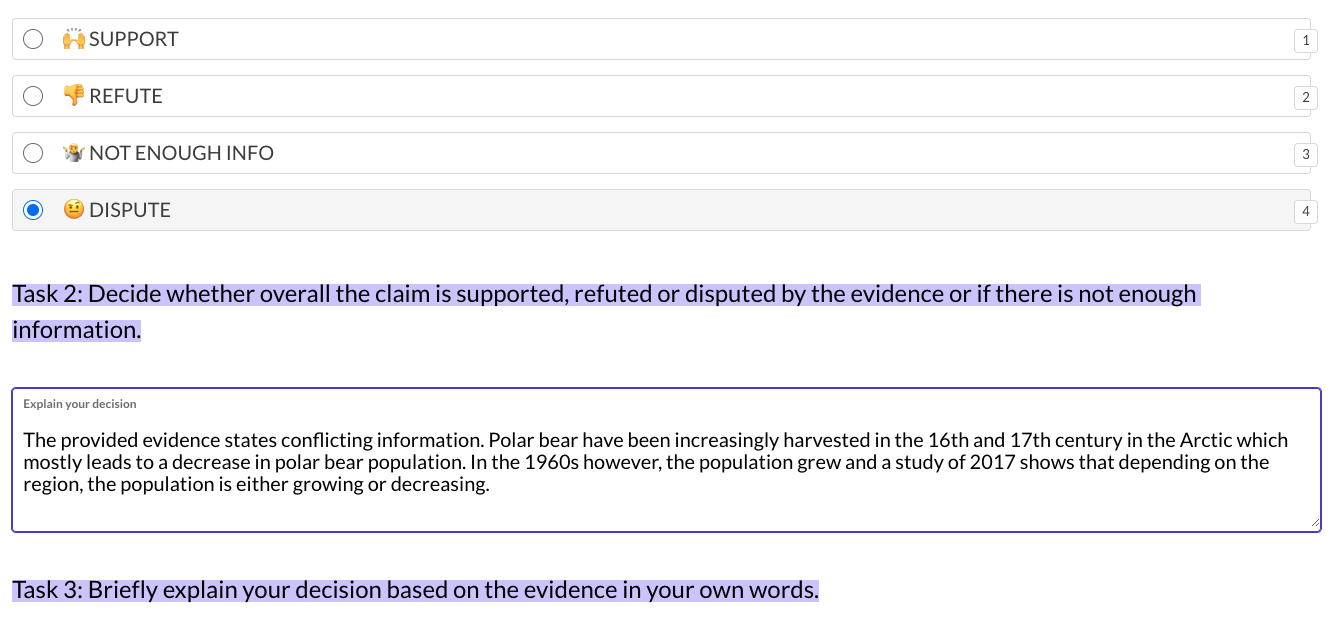}
    \caption{Prodigy annotation framework for one claim. Parts of the evidences are annotated as either supporting (red) or contradicting (blue).}
    \label{fig:polarbear}
\end{figure*}

Annotators were asked to solve 4 tasks, an example is shown in Figure \ref{fig:polarbear}.
\begin{enumerate}[noitemsep]
\setcounter{enumi}{-1}
    \item reading the claim and all 5 evidences
    \item token-level rationale annotation for each \\evidence statement
    \item decide on a claim label
    \item briefly explain the decision
\end{enumerate}
We publish the entire study but will focus on rationale annotation and claim labels in this paper.
\begin{figure}
    \centering
    \includegraphics[width=0.75\linewidth]{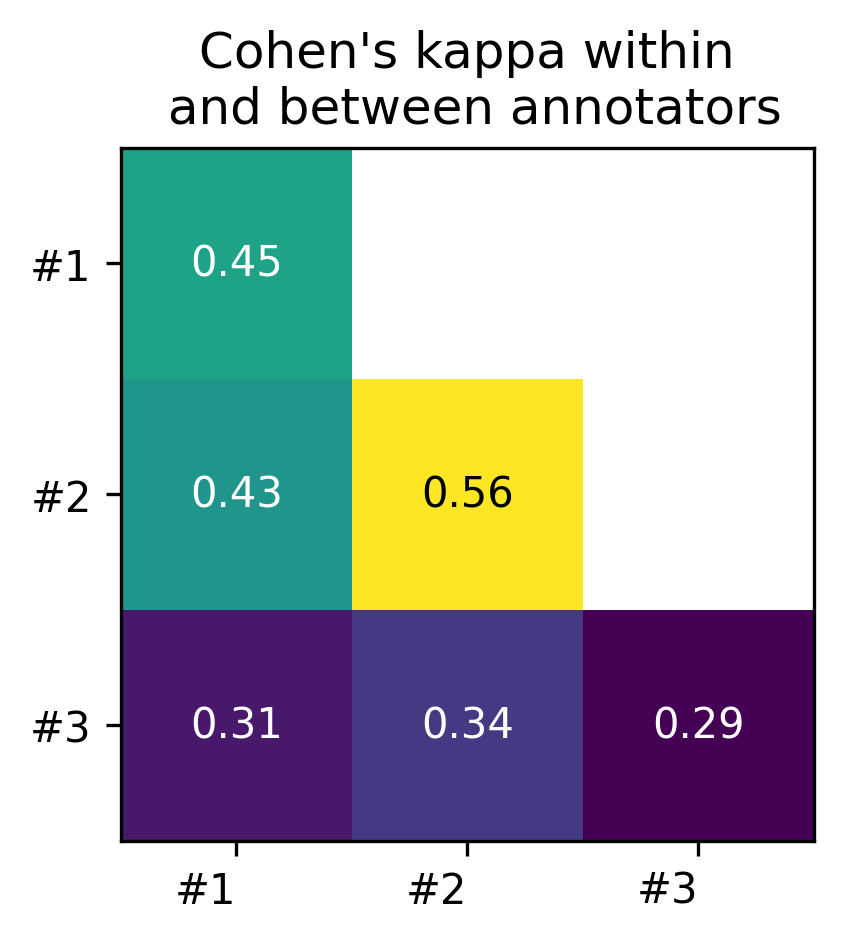}
    \caption{Kappa scores for inter and intra annotator agreement of all 3 annotators.}
    \label{fig:IOA}
\end{figure}
\subsection{First and last batch}
104 samples were divided into 10 batches with 12 samples each, except for the last batch which only contained 6 samples and a pre-trial which contained 3 samples. Since batches were annotated across several weeks, we repeated the 3 samples from the pre-trial to the last batch in order to compute intra-annotator agreements on those 3 samples.
\subsection{Inter- and Intra- Annotator Agreements}
We show Cohen's kappa scores for inter- and intra-annotator agreement in Figure \ref{fig:IOA}. For the inter-annotator agreement, we concatenated all annotations and computed pairwise kappa scores. For the intra-annotator agreement, we computed scores on the overlap between pre-trial and the last batch for each annotator. Average inter annotator agreement is $0.36\pm 0.05$ (kappa) and $0.57 \pm 0.04$ (macro F1).
\subsection{Comparison to previous annotations}
We compare labels on the newly annotated subset of Climate-Fever with previously annotated labels. There are several differences in the annotation process that require attention when interpreting the results. In the original study, evidences were labeled by multiple annotators and claim labels then inferred from two stages of majority voting (see previous section). In our annotation study, we annotated evidences with rationales and inferred evidence labels based on supporting evidence (\textsc{support}), contradicting evidence (\textsc{refute}), no evidence (\textsc{not enough information}) or disputing evidence (\textsc{dispute}). Claim labels were annotated after the rationales annotation. This also means that in our annotation study, evidences could get a \textsc{dispute} label which was not the case in the original data collection process. Tables \ref{tab:claimf1} and \ref{tab:evidence1} show classification results where \textit{new} labels are considered as the ground truth. Results show that label agreement on the claim level is much higher for \textsc{support} and \textsc{refute} than for the other two labels. On the evidence level we see balanced F1 scores around $0.6 - 0.66$ for all overlapping labels but a score of $0$ for the \textsc{dispute} class that only exist in the new annotations. Those scores are comparable to the previously reported average kappa and macro-F1 scores for the inter- and intra-annotator agreements.
\begin{table}[]
    \centering
    \begin{tabular}{r|c|c|c}
                        & Prec. & Recall & F1-score \\ \hline
       \textsc{Support}         & 0.75  & 0.83   &   0.79\\
       \textsc{Refute}          & 0.71  & 0.87   &   0.78\\
       \textsc{Not enough info} & 0.00  & 0.00   &   0.00 \\
       \textsc{Dispute}         & 0.46  & 0.43   &   0.44 \\ \hline \hline
       Macro-F1                 &       &        & 0.50
    \end{tabular}
    \caption{Comparison between original claim labels and newly annotated claim labels}
    \label{tab:claimf1}
\end{table}

\begin{table}[]
    \centering
    \begin{tabular}{r|c|c|c}
                        & Prec. & Recall & F1-score \\ \hline
       \textsc{Support}         & 0.56  & 0.79   &   0.65\\
       \textsc{Refute}          & 0.54  & 0.67   &   0.60\\
       \textsc{Not enough info} & 0.74  & 0.60   &   0.66 \\
       \textsc{Dispute}         & 0.00  & 0.00   &   0.00 \\ \hline \hline
       Macro-F1                 &       &        & 0.48
    \end{tabular}
    \caption{Comparison between original evidence labels and newly annotated evidence labels}
    \label{tab:evidence1}
\end{table}

\section{Model experiments}\label{sec:models}
\subsection{Models}
For our experiments, we use the following instruction-tuned LLMs: Gemma3-12b\footnote{\href{https://huggingface.co/google/gemma-3-12b-it}{\nolinkurl{huggingface.co/google/gemma-3-12b-it}}}, Llama3.1-8B\footnote{\href{https://huggingface.co/meta-llama/Meta-Llama-3.1-8B-Instruct}{\nolinkurl{huggingface.co/meta-llama/Meta-Llama-3.1-8B-Instruct}}}, Qwen3-8B\footnote{\href{https://huggingface.co/Qwen/Qwen3-8B}{\nolinkurl{https://huggingface.co/Qwen/Qwen3-8B}}}, Mistral-7B\footnote{\href{https://huggingface.co/mistralai/Mistral-7B-Instruct-v0.3}{\nolinkurl{https://huggingface.co/mistralai/Mistral-7B-Instruct-v0.3}}}
Although some of them might not be considered large in size, we follow the convention of calling them Large Language Models based on their abilities and training procedures.


\subsection{Instructions} \label{app:instruct}
\begin{figure}[h!]
\centering
    \includegraphics[width=0.5\textwidth]{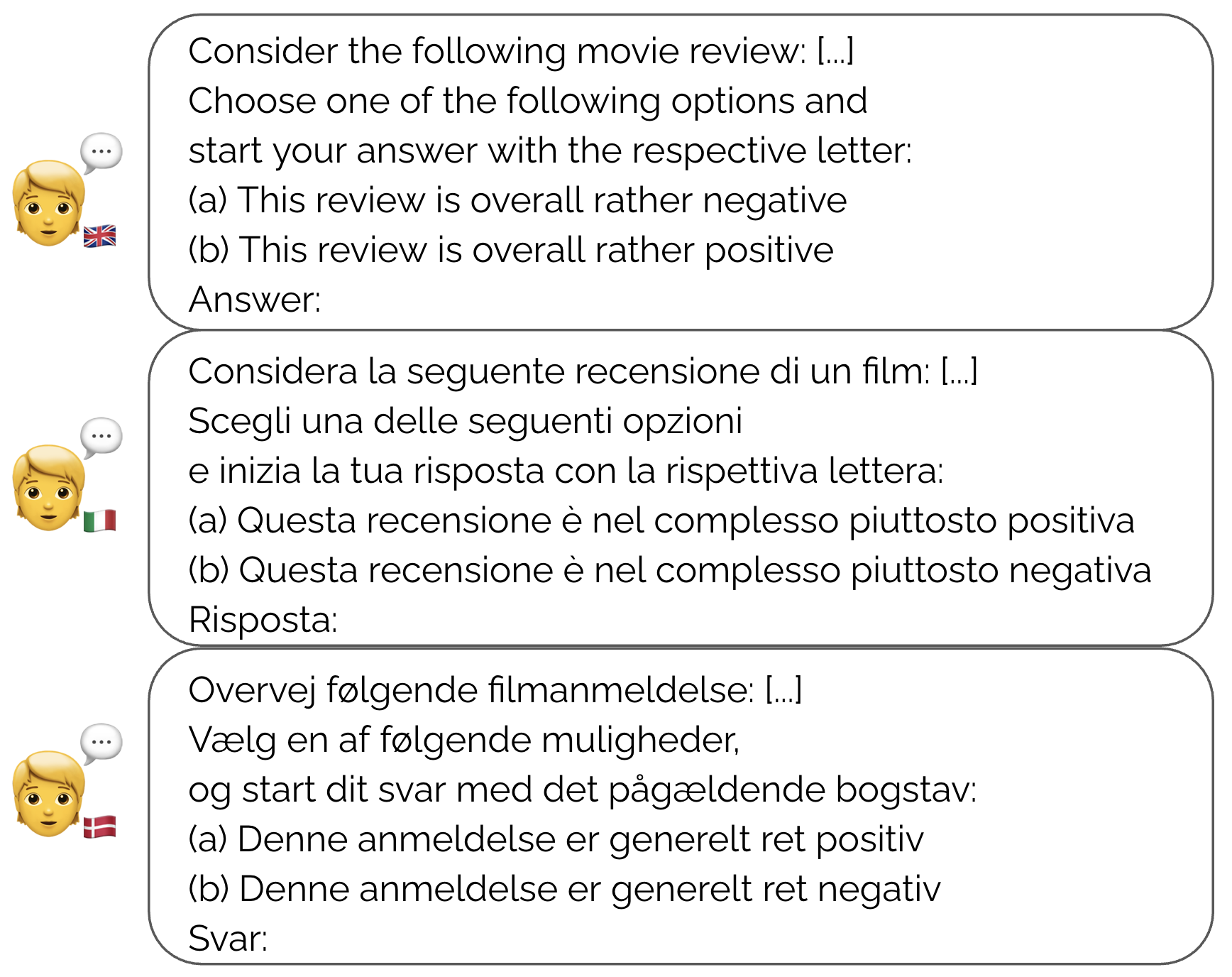}
    \caption{Prompts in all 3 languages to solve sentiment classification.}
    \label{fig:prompts_sst}
\end{figure}

\begin{figure}[h!]
\centering
    \includegraphics[width=0.5\textwidth]{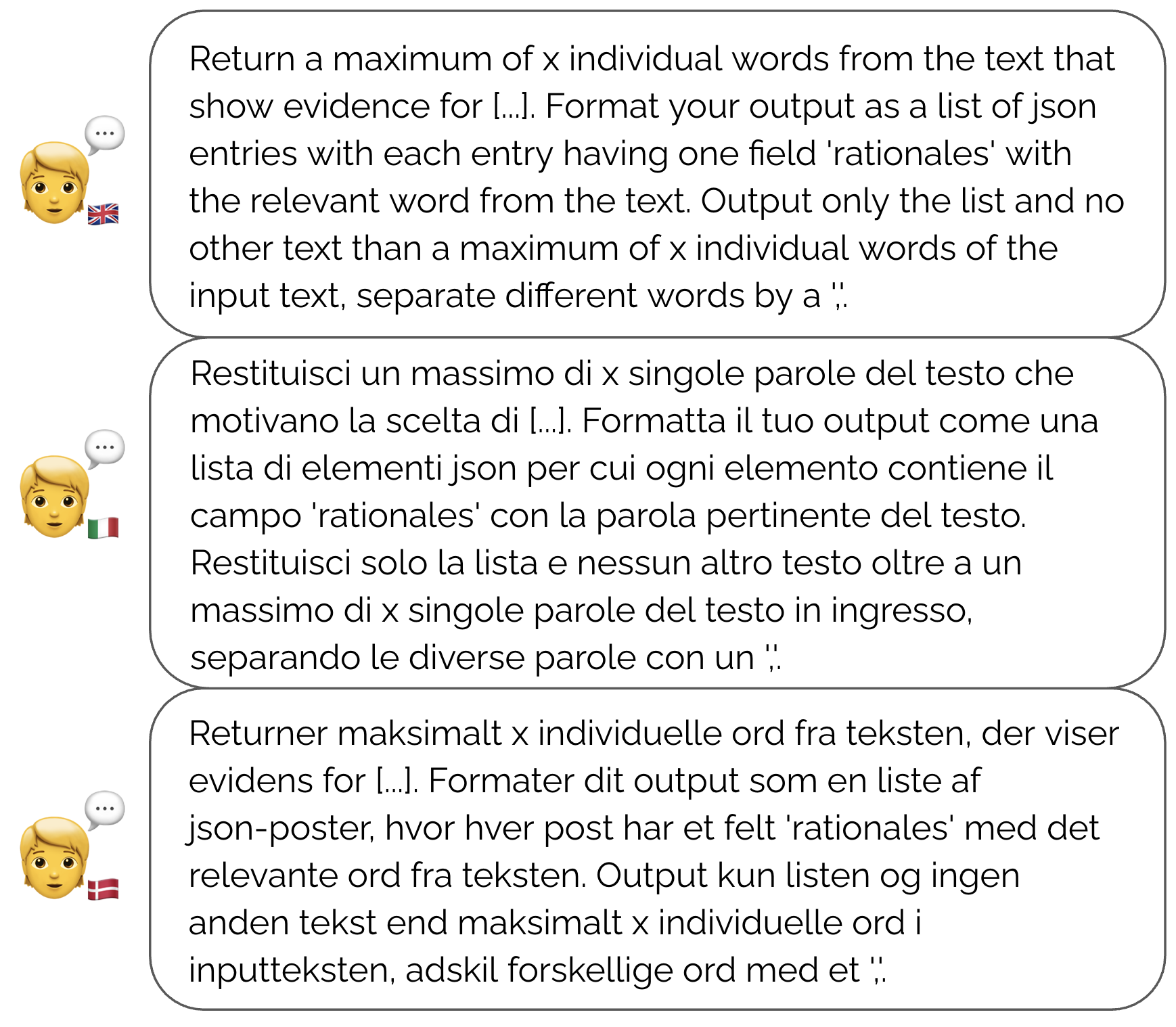}
    \caption{Follow-up prompts in all 3 languages to extract rationales.}
    \label{fig:follow-up_sst}
\end{figure}

\begin{figure}[h!]
\centering
    \includegraphics[width=0.5\textwidth]{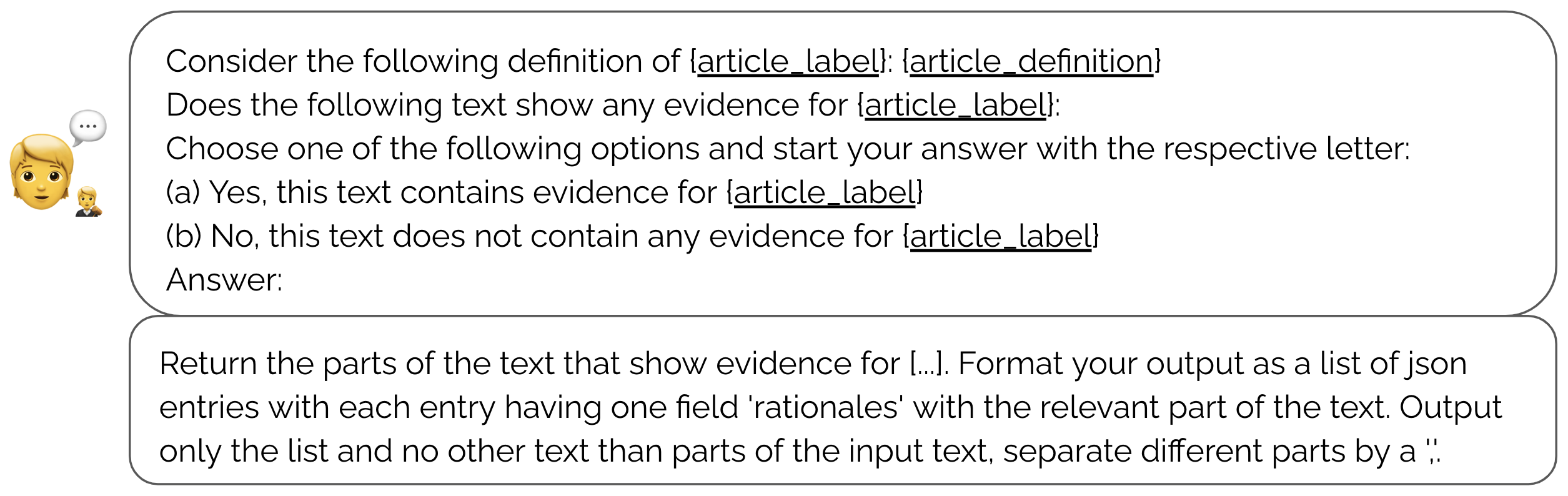}
    \caption{Prompts for classification and rationale extraction for the RaFoLa dataset.}
    \label{fig:rafola_instructions}
\end{figure}

\begin{figure}[h!]
\centering
    \includegraphics[width=0.5\textwidth]{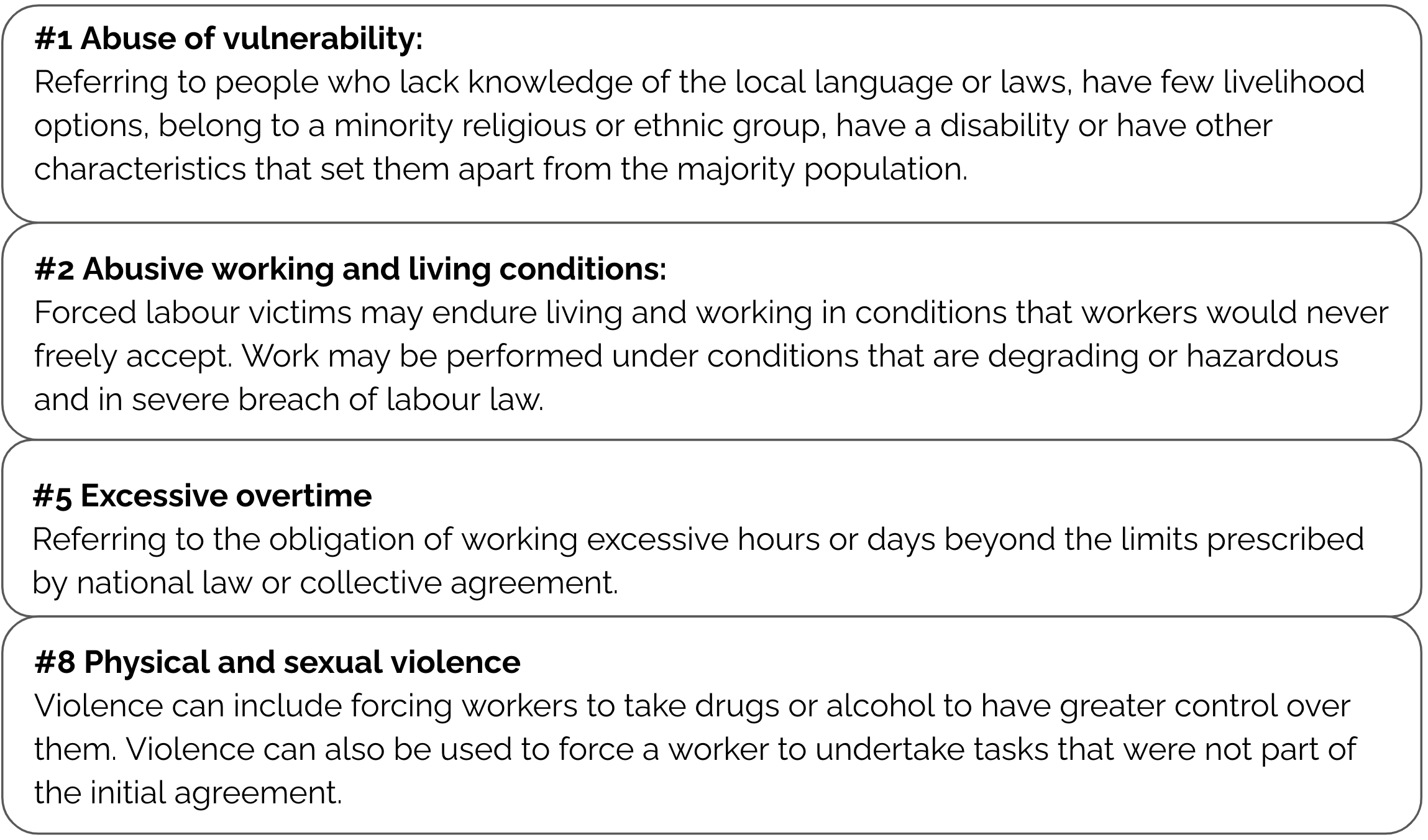}
    \caption{Indicators defined by the International Labour Organization and published by \citeauthor{mendez-guzman-etal-2022-rafola}.}
    \label{fig:rafola_definitions}
\end{figure}

We show multilingual instructions for SST in Figures \ref{fig:prompts_sst} - \ref{fig:follow-up_sst}. Instructions for forced labour detection in RaFoLa are shown in Figure \ref{fig:rafola_instructions} and relevant definitions from the International Labour Organization in Figure \ref{fig:rafola_definitions}.


\section{Corpus Statistics \& Top-8 tokens}\label{app:corpus_stats}
\begin{table*}[h]
\setlength{\tabcolsep}{2pt} 

\scriptsize
\caption{Corpus Statistics across all three datasets. Abbreviations as follow: MDD: Mean dependency depth (syntactic complexity), TTR: Token type ratio, POS: Fraction of pos entities, GPE: Geopolitical entity, ORG: Organization entity, NORP: Nationalities or religious or political groups.}

\label{tab:corpus_stats}
\centering

\begin{tabular}{lrrrrrrrrrrrrrr}
\toprule
\multirow{ 2}{*}{Dataset} & \multirow{ 2}{*}{Toks/Doc} & \multirow{ 2}{*}{Toks/Sent} & \multirow{ 2}{*}{MDD} & TTR  & Stopwords  & Formatting  & POS  & Lex. Ambig.  & GPE & Cardinal  & ORG & Date & NORP  & Person \\
&&&&[\%]&[\%]&[\%]&[\%]&[\%]&[\%]&[\%]&[\%]&[\%]&[\%]&[\%]\\
\midrule
SST & 20.86 & 20.71 & 2.75 & 36.78 & 43.03 & 12.65 & 3.52 & 54.07 & 0.36 & 0.35 & 0.67 & 0.40 & 0.24 & 1.49 \\
RaFoLa & 944.89 & 29.64 & 2.92 & 3.30 & 38.37 & 11.82 & 6.47 & 54.03 & 1.43 & 0.80 & 1.90 & 0.98 & 0.47 & 0.89 \\
Climate-Fever & 199.86 & 16.70 & 1.87 & 4.88 & 33.33 & 14.42 & 7.22 & 54.10 & 0.47 & 3.60 & 0.96 & 1.74 & 0.13 & 0.32 \\
\bottomrule
\hspace{2cm}
\end{tabular}
\end{table*}

\begin{table}[h!]
\begin{tabular}{p{1.2cm}|p{6cm}}
\toprule
corpus & climate, global, ice, sea, warming, change, greenhouse, human \\
\midrule
human & global, sea, climate, ice, warming, earth, greenhouse, human \\
\midrule
gemma3 & climate, global, greenhouse, change, warming, sea, ice, temperatures \\
\midrule
llama3 & global, climate, warming, ice, change, sea, temperatures, greenhouse \\
\midrule
qwen3 & ice, sea, climate, warming, global, level, temperature, mass \\
\midrule
mistral & ice, global, warming, heat, temperatures, greenhouse, human, increased \\
\bottomrule
\end{tabular}
\caption{Top-8 tokens from Climate-Fever (first row) and from respective rationale annotations by humans (2nd row) and models.}
\label{tab:top8-climate}
\end{table}

\begin{table*}
\small
\begin{tabular}{p{1.2cm} |  p{3cm} | p{3cm} | p{3.5cm} | p{3.5cm} }
\toprule
 & SST &  mSST-EN &  mSST-DA &  mSST-IT \\
\midrule
corpus & movie, film, like, comedy, work, -, love, funny & film, movie, performances, characters, bad, funny, like, story & film, `, filmen, ', sjov, karakterer, bare, filmens & film, i, divertente, personaggi, interpretazioni, storia, trama, avvincente \\
\midrule
human & funny, best, bad, movie, beautifully, compelling, film, performance & performances, bad, funny, good, characters, dull, film, compelling & sjov, film, overbevisende, præstationer, bedste, plot, vittig, sjovt & divertente, avvincente, noioso, interpretazioni, film, ben, brutto, assolutamente \\
\midrule
gemma3 & best, funny, bad, love, beautifully, compelling, hilarious, fun & funny, bad, performances, dull, compelling, good, long, best & sjov, overbevisende, spændende, humor, tilfredsstillende, klodset, dårlig, dårligt & divertente, avvincente, film, noioso, ben, brutto, intelligente, umorismo \\
\midrule
llama3 & best, funny, bad, beautifully, year, little, compelling, hilarious & bad, performances, funny, good, dull, best, compelling, little & sjov, overbevisende, bare, dårlig, dårligt, kedelig, spænding, spændende & divertente, avvincente, umorismo, noioso, senso, ben, interpretazioni, intelligente \\
\midrule
qwen3 & bad, best, love, movie, funny, beautifully, compelling, film & funny, bad, performances, dull, compelling, good, comedy, little & sjov, spændende, præstationer, overbevisende, vittig, sjovt, komedie, dårlig & divertente, avvincente, film, noioso, interpretazioni, personaggi, intelligente, intelligenti \\
\midrule
mistral & comedy, best, bad, beautifully, funny, little, fun, stupid & bad, performances, funny, dull, characters, best, film, intelligent & sjov, filmen, præstationer, plot, film, spændende, sentimentalitet, giver & divertente, film, interpretazioni, i, trama, personaggi, avvincente, ben \\
\bottomrule
\end{tabular}
\caption{Top-8 tokens from SST/mSST splits (first row) and from respective rationale annotations by humans (2nd row) and models.}
\label{tab:top8-SST}
\end{table*}

\begin{table*}
\small
\begin{tabular}{p{1.2cm} | p{3.2cm} | p{3.2cm} | p{3.2cm} | p{3.2cm} }
\toprule
 &  \#1 Abuse of vulnerability &  \#2 Abusive working and living conditions  &  \#5 Excessive overtime & \#8 Physical and sexual violence \\
\midrule
corpus & \multicolumn{4}{l}{\makecell[l]{workers, said, labour, work, rights, labor, human, children}}\\
\midrule
human & work, workers, children, forced, women, labour, said, vulnerable & workers, conditions, work, little, water, working, forced, said & hours, day, working, 12, work, worked, days, week & abuse, sexual, harassment, said, physical, women, violence, verbal \\
\midrule
gemma3 & workers, work, labour, forced, said, children, women, rights & workers, work, forced, conditions, labour, said, working, children & hours, work, day, days, working, workers, forced, week & workers, sexual, abuse, said, women, forced, violence, harassment \\
\midrule
llama3 & workers, work, said, labour, forced, children, women, working & workers, work, said, working, conditions, forced, labour, day & hours, day, working, said, work, workers, 12, days & said, sexual, workers, abuse, women, harassment, violence, work \\
\midrule
qwen3 & workers, work, forced, labour, said, children, women, working & workers, work, said, forced, labour, conditions, working, children & work, hours, workers, said, day, working, labour, forced & workers, said, abuse, sexual, forced, women, work, violence \\
\midrule
mistral & workers, work, said, forced, conditions, working, children, women & workers, work, said, conditions, forced, labour, working, day & day, contracts, working, hours, work, said, days, supervisors & said, sexual, women, factory, abuse, woman, report, physical \\
\bottomrule
\end{tabular}
\caption{Top-8 tokens from RaFoLa (first row) and from respective rationale annotations by humans (2nd row) and models.}
\label{tab:top8-Rafola}
\end{table*}

\newpage
\section{Faithful rationale comparison}\label{app:faithful_rationales}

\subsection{Methodological Details} 
We compare human and model-based rationales provided on the same samples from various annotation studies and evaluate plausibility, i.e., agreement with human rationales and faithfulness to the model in comparison to state-of-the-art gradient-based feature attribution methods such as layer-wise relevance propagation (LRP)\cite{transformerxai2022} and GradientxInput  across models. We provide details on the LRP propagation procedure in the following.
 \paragraph{LRP}

We applied the LN-rule \cite{transformerxai2022} for LayerNorm/RMSNorm, the Identity-rule \cite{NEURIPS2024_d6d0e41e} for nonlinear activation functions, and the LRP-$0$ rule \cite{montavon2019layer} for linear transformations. The AH-rule was used for Mistral as it yielded better faithfulness compared to application of the Half-rule \cite{NEURIPS2024_d6d0e41e, Arras2019} (see \cite{NEURIPS2024_d6d0e41e, jafari2025relp} for further discussion). A summary of implemented rules used in our experiments is given in Table~\ref{tab:lrp_details}.

\begin{table}[tbh]
\centering
\renewcommand{\arraystretch}{1.1} 
\footnotesize
\resizebox{\linewidth}{!}{
\begin{tabular}{lcccc}
\toprule
\textbf{Propagation Rule} & 
\rotatebox{90}{\textbf{Mistral}} & 
\rotatebox{90}{\textbf{Llama3}} & 
\rotatebox{90}{\textbf{Qwen3}} & 
\rotatebox{90}{\textbf{Gemma3}} \\
\midrule
LN-rule \cite{transformerxai2022} & \checkmark & \checkmark & \checkmark & \checkmark \\
Identity-rule \cite{NEURIPS2024_d6d0e41e} & \checkmark & \checkmark & \checkmark & \checkmark \\
0-rule \cite{montavon2019layer} & \checkmark & \checkmark & \checkmark & \checkmark \\
Half-rule \cite{NEURIPS2024_d6d0e41e} & \texttimes & \checkmark & \checkmark & \checkmark \\
AH-rule \cite{transformerxai2022} & \checkmark & \texttimes & \texttimes & \texttimes \\
\bottomrule
\end{tabular}
}
\caption{Propagation rules used in our LRP implementation across model families. A checkmark (\checkmark) indicates the rule was applied, and a cross (\texttimes) indicates it was not.}
\label{tab:lrp_details}
\end{table}

\subsection{LLM-assisted Analysis of Rationale Tokens} \label{app:llm_analysis}
We use GPT-5 for an initial analysis of the large number of faithful token subsets derived from human rationales, model-based self-explanations, and LRP post-hoc explanations. Specifically, we prompted GPT-5 with: “Given token lists for multiple samples, summarize the main similarities and differences across Groups H, M, and P.”. This task prompt was followed by up to 50 randomly selected samples per group (human, model, post-hoc), each containing the top 5\% of tokens. A randomly selected subset of token lists is also presented in Tables \ref{tab:rafola_samples_art1} and \ref{tab:rafola_samples_art8}. All text inputs used for the LLM-assisted analysis will be made available upon publication.
This procedure was repeated for each model and used for hypothesis generation and initial exploratory corpus analysis. We then validated this qualitative analysis with statistical methods and standard automated NLP pipelines, including entity identification, stopword filtering, and lexical diversity assessment as presented in Table \ref{tab:faithful_rationales}.

\begin{figure*}[ht!]
    \includegraphics[width=\textwidth, trim={0cm 1.1cm 0cm 0.6cm},
  clip]{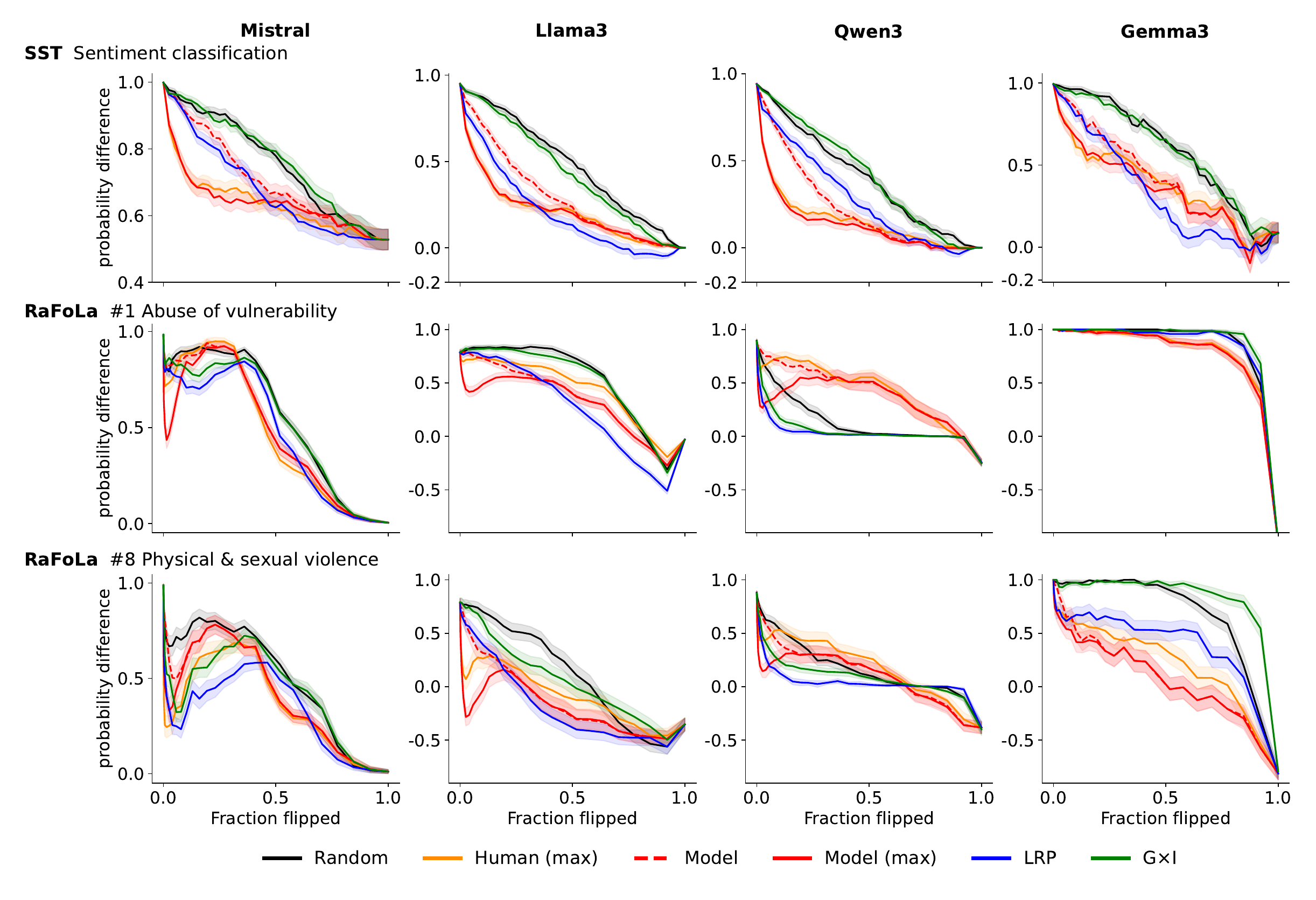}
    \caption{Faithfulness evaluation for SST and RaFoLa (articles \#1 and \#8). Model probability difference after masking tokens extracted from human rationales, model self-explanation rationales and post-hoc attributions (LRP, GxI) across models. Shaded bands indicate standard errors across samples. Faster drop in probability for early fractions indicates more faithful identification of task-relevant rationales. Human/Model (max) refers to rationales selected via greedy maximization of next-token probability difference.}
    \label{fig:faithfulness_app} 
    \vspace{-5pt}
\end{figure*}

\begin{table*}[h]
\footnotesize
\caption{Token-level statistics for different models and sources. Numeric values are percentages (\%). TTR refers to Type-Token ratios, measuring lexical diversity, and entity columns correspond to the following semantic categories: GPE (countries, cities, locations), Cardinal (numbers), ORG (organizations), Date (temporal expressions), NORP (nationalities, religious or political groups), and PERSON (individuals). Human, model, and post-hoc rows correspond to different rationale sources, with $\Delta$ rows showing differences relative to the human baseline.}

\label{tab:faithful_rationales}
\centering
\setlength{\tabcolsep}{1.8pt} 

\begin{tabular}{%
  >{\raggedright\arraybackslash}p{1.2cm} 
  >{\raggedright\arraybackslash}p{1.4cm} 
  *{9}{>{\raggedleft\arraybackslash}p{1.2cm}} 
}
\toprule
\multicolumn{1}{c}{Model} & 
\multicolumn{1}{c}{Source} & 
\multicolumn{1}{c}{\hspace{10pt} TTR} & 
\multicolumn{1}{c}{Stopwords} & 
\multicolumn{1}{c}{\hspace{3pt} Formatting} & 
\multicolumn{1}{c}{\hspace{10pt} GPE} & 
\multicolumn{1}{c}{\hspace{4pt} Cardinal} & 
\multicolumn{1}{c}{\hspace{11pt} ORG} & 
\multicolumn{1}{c}{\hspace{9pt} Date} & 
\multicolumn{1}{c}{\hspace{6pt} NORP} & 
\multicolumn{1}{c}{\hspace{3pt} Person} \\
\midrule
\multicolumn{11}{c}{RaFoLa: \# 1} \\
\midrule
\midrule
Llama3 & human & 31.91 & 38.35 & 2.87 & 0.89 & 1.15 & 0.71 & 0.52 & 0.64 & 0.92 \\
Llama3 & model & 34.43 & 35.28 & 3.25 & 1.04 & 1.34 & 1.39 & 0.54 & 0.87 & 0.95 \\
Llama3 & post-hoc & 31.87 & 14.02 & 17.44 & 1.69 & 2.15 & 2.22 & 0.33 & 0.73 & 2.23 \\
\midrule

Mistral & human & 32.16 & 36.41 & 2.35 & 0.50 & 1.08 & 0.87 & 0.48 & 0.46 & 0.71 \\
Mistral & model & 32.05 & 37.23 & 1.66 & 0.48 & 1.45 & 0.87 & 0.46 & 0.35 & 0.64 \\
Mistral & post-hoc & 25.11 & 12.57 & 31.97 & 0.56 & 2.06 & 0.98 & 0.54 & 0.25 & 0.91 \\
\midrule
Qwen3 & human & 44.43 & 37.92 & 2.78 & 0.62 & 1.36 & 0.90 & 0.51 & 0.73 & 0.34 \\
Qwen3 & model & 47.33 & 38.42 & 2.34 & 1.01 & 1.57 & 1.35 & 0.45 & 0.51 & 1.12 \\
Qwen3 & post-hoc & 37.47 & 19.60 & 16.98 & 1.07 & 0.95 & 2.74 & 0.36 & 0.83 & 1.90 \\
\midrule

Gemma3 & human & 29.19 & 38.03 & 1.79 & 0.87 & 1.27 & 0.53 & 0.41 & 0.73 & 0.75 \\
Gemma3 & model & 28.68 & 37.36 & 1.35 & 1.01 & 1.37 & 0.80 & 0.60 & 0.70 & 0.63 \\
Gemma3 & post-hoc & 24.18 & 25.56 & 6.19 & 0.87 & 2.06 & 1.82 & 0.63 & 0.57 & 1.37 \\

\midrule
\midrule

Llama3 & \multirow{4}{*}{\rotatebox{90}{\parbox{1.4cm}{\centering $\Delta$ human - model}}} & -2.52 & 3.07 & -0.38 & -0.15 & -0.19 & -0.68 & -0.02 & -0.23 & -0.03 \\
Mistral &  & 0.11 & -0.82 & 0.69 & 0.02 & -0.37 & 0.00 & 0.02 & 0.11 & 0.07 \\
Qwen3 & & -2.90 & -0.50 & 0.44 & -0.39 & -0.21 & -0.45 & 0.06 & 0.22 & -0.78 \\
Gemma3 &  & 0.51 & 0.67 & 0.44 & -0.14 & -0.10 & -0.27 & -0.19 & 0.03 & 0.12 \\
\midrule
Llama3 & \multirow{4}{*}{\rotatebox{90}{\parbox{1.4cm}{\centering $\Delta$ human - post-hoc}}}& 0.04 & 24.33 & -14.57 & -0.80 & -1.00 & -1.51 & 0.19 & -0.09 & -1.31 \\
Mistral &  & 7.05 & 23.84 & -29.62 & -0.06 & -0.98 & -0.11 & -0.06 & 0.21 & -0.20 \\
Qwen3 & & 6.96 & 18.32 & -14.20 & -0.45 & 0.41 & -1.84 & 0.15 & -0.10 & -1.56 \\
Gemma3 & & 5.01 & 12.47 & -4.40 & 0.00 & -0.79 & -1.29 & -0.22 & 0.16 & -0.62 \\

\bottomrule

\midrule
\multicolumn{11}{c}{RaFoLa: \# 8} \\

\midrule
\midrule
Llama3 & human & 32.93 & 38.57 & 2.93 & 0.36 & 0.97 & 0.69 & 0.40 & 0.47 & 0.97 \\
Llama3 & model & 35.07 & 38.85 & 3.18 & 0.47 & 0.87 & 0.98 & 0.47 & 0.54 & 0.83 \\
Llama3 & post-hoc & 34.36 & 15.70 & 16.80 & 1.25 & 1.33 & 2.00 & 0.94 & 0.43 & 2.63 \\
\midrule
Mistral & human & 31.40 & 34.99 & 2.29 & 0.24 & 0.87 & 1.02 & 0.35 & 0.24 & 1.06 \\
Mistral & model & 33.18 & 37.63 & 2.68 & 0.28 & 0.91 & 1.38 & 0.39 & 0.12 & 0.99 \\
Mistral & post-hoc & 25.65 & 13.63 & 32.94 & 0.83 & 1.22 & 0.91 & 0.63 & 0.28 & 0.87 \\
\midrule

Qwen3 & human & 40.80 & 38.76 & 2.33 & 0.59 & 1.72 & 1.08 & 0.34 & 0.49 & 0.69 \\
Qwen3 & model & 40.70 & 38.85 & 3.40 & 0.59 & 2.21 & 0.79 & 0.39 & 0.49 & 0.69 \\
Qwen3 & post-hoc & 32.98 & 17.97 & 17.24 & 1.20 & 0.89 & 1.99 & 0.58 & 0.84 & 2.09 \\
\midrule
Gemma3 & human & 29.79 & 36.07 & 2.05 & 0.66 & 1.21 & 0.80 & 0.25 & 0.73 & 0.93 \\
Gemma3 & model & 30.38 & 37.14 & 1.96 & 0.82 & 1.12 & 0.80 & 0.59 & 0.77 & 0.64 \\
Gemma3 & post-hoc & 25.42 & 23.37 & 6.55 & 0.73 & 2.44 & 1.41 & 0.84 & 0.25 & 1.25 \\

\midrule
\midrule

Llama3 & \multirow{4}{*}{\rotatebox{90}{\parbox{1.4cm}{\centering $\Delta$ human - model}}}  & -2.14 & -0.28 & -0.25 & -0.11 & 0.10 & -0.29 & -0.07 & -0.07 & 0.14 \\
Mistral  &      & -1.78 & -2.64 & -0.39 & -0.04 & -0.04 & -0.36 & -0.04 & 0.12 & 0.07 \\
Qwen3 &       & 0.10 & -0.09 & -1.07 & 0.00 & -0.49 & 0.29 & -0.05 & 0.00 & 0.00 \\
Gemma3 &       & -0.59 & -1.07 & 0.09 & -0.16 & 0.09 & 0.00 & -0.34 & -0.04 & 0.29 \\
\midrule

Llama3 &  \multirow{4}{*}{\rotatebox{90}{\parbox{1.4cm}{\centering $\Delta$ human - post-hoc}}} & -1.43 & 22.87 & -13.87 & -0.89 & -0.36 & -1.31 & -0.54 & 0.04 & -1.66 \\
Mistral &  & 5.75 & 21.36 & -30.65 & -0.59 & -0.35 & 0.11 & -0.28 & -0.04 & 0.19 \\
Qwen3 & & 7.82 & 20.79 & -14.91 & -0.61 & 0.83 & -0.91 & -0.24 & -0.35 & -1.40 \\
Gemma3 &  & 4.37 & 12.70 & -4.50 & -0.07 & -1.23 & -0.61 & -0.59 & 0.48 & -0.32 \\
\bottomrule
\end{tabular}

\end{table*}

\begin{table*}[h]
\footnotesize
\caption{Qualitative examples of extracted tokens across models (rows) and  human-annotated rationales, model-generated rationales, and post-hoc attribution-based (LRP) rationale for the RaFoLa dataset (here: article \#1). Tokens are selected from the top 5\% most faithful tokens in randomly selected samples. Irregular spacing can occur due to differences in tokenizers’ subtoken processing strategies.}

\label{tab:rafola_samples_art1}
\centering
\setlength{\tabcolsep}{1.8pt} 

\begin{tabular}{l p{5.5cm} p{5.5cm} p{3.6cm}}
\toprule
model & human & model & post-hoc \\
\midrule
\midrule
Llama3 & \begin{itemize}[leftmargin=*, label=-, labelsep=0.3em, itemsep=0pt, topsep=0pt, partopsep=0pt, parsep=0pt]
\item fearful Workers looking fearful could be a sign
\item with He said that in the past two years victims of trafficking had been found in hotels in North Wales where organised crime gangs had set up
\item serv itude . When choosing their victims traff ickers target the most vulnerable
\item monitoring to ensure children and other vulnerable groups
\item People working as cooks bus staff and wait staff might be exploited with traff ickers often taking advantage of language barriers between exploited workers and patrons
\end{itemize} & \begin{itemize}[leftmargin=*, label=-, labelsep=0.3em, itemsep=0pt, topsep=0pt, partopsep=0pt, parsep=0pt]
\item million labour ers the Uzbek harvest is the biggest recruitment programme anywhere in the world according to the International Labor Organization I LO ). U zbek istan
\item and girls are disproportionately affected accounting
\item Fatal accidents and child labour are common
\item from China and South Korea .
\item every where . X in jiang the north western region of China is home to minority
\end{itemize} & \begin{itemize}[leftmargin=*, label=-, labelsep=0.3em, itemsep=0pt, topsep=0pt, partopsep=0pt, parsep=0pt]
\item Three
\item .aspx
\item forced
\item Swiss -backed
\item BBC
\item UK
\item Officers
\item Gang
\item appeal
\item Wales
\item modern slavery
\item trafficking
\end{itemize} \\
\midrule
Mistral & \begin{itemize}[leftmargin=*, label=-, labelsep=0.3em, itemsep=0pt, topsep=0pt, partopsep=0pt, parsep=0pt]
\item Work ers looking fear ful could be a sign of
\item He said that in the past two years victims of traff icking had been found in hotels in North Wales where organ ised crime gang s had set up
\item Bul gar ian nation als from disadv
\item When choosing their victims traff ick ers target the most vulnerable
\item The U igh urs had to come because they
\end{itemize} & \begin{itemize}[leftmargin=*, label=-, labelsep=0.3em, itemsep=0pt, topsep=0pt, partopsep=0pt, parsep=0pt]
\item Work ers looking fear ful could be a sign of
\item I remember thinking What is it that they have ? Because I needed that in my own
\item Work ers in the Indian capital build buildings for about
\item ly co er ced to work against their will were members of the Muslim U igh ur minority in north western
\item t ann ery workers do not have
\end{itemize} & \begin{itemize}[leftmargin=*, label=-, labelsep=0.3em, itemsep=0pt, topsep=0pt, partopsep=0pt, parsep=0pt]
\item Modern
\item S la
\item published
\item 0x0A
\item An appeal has
\item rural businesses
\item gang
\item modern slavery
\item Bulgar
\item '“'
\item http
\item 0x0A Acc ording
\end{itemize} \\
\midrule
Qwen3 & \begin{itemize}[leftmargin=*, label=-, labelsep=0.3em, itemsep=0pt, topsep=0pt, partopsep=0pt, parsep=0pt]
\item serv itude . When choosing their victims traff ickers target the most vulnerable
\item foreign workers most of whom are women have very little means to defend themselves should the employer abuse them in any way
\item others . "One year I was pregnant but it
\item If they are undocumented migrants they may also fear that coming forward will result in their deportation . They
\item Some never go to school or learn to read and write Others are smugg led across borders and left vulnerable to trafficking or sexual abuse . The
\end{itemize} & \begin{itemize}[leftmargin=*, label=-, labelsep=0.3em, itemsep=0pt, topsep=0pt, partopsep=0pt, parsep=0pt]
\item Forum for Human Rights reported that
\item serv itude . When choosing their victims traff ickers target
\item ok hid akh on K has anova was signed out
\item false job prospects and cash loans with ex orbit ant interest rates
\item ranks of workers without social protections including low wages and employer controls
\end{itemize} & \begin{itemize}[leftmargin=*, label=-, labelsep=0.3em, itemsep=0pt, topsep=0pt, partopsep=0pt, parsep=0pt]
\item Swiss
\item programme
\item female
\item In the
\item speak
\item photographer Lisa Krist
\item Lisa Krist
\item Disclaimer
\item film director
\item Twitter
\item By
\item exploitation
\end{itemize} \\
\midrule
Gemma3 & \begin{itemize}[leftmargin=*, label=-, labelsep=0.3em, itemsep=0pt, topsep=0pt, partopsep=0pt, parsep=0pt]
\item In the past schoolchildren university students doctors teachers and other public sector professionals were required to participate in the harvest for little or no pay and under threat of punishment
\item fearful Workers looking fearful could be a sign of the various methods used to control workers who are being exploited
\item with .” He said that in the past two years victims of trafficking had been found in hotels in North Wales where organised crime gangs had set up
\item network Bul garian nationals from disadvantaged regions were recruited by
\item servitude When choosing their victims traffickers target the most vulnerable
\end{itemize} & \begin{itemize}[leftmargin=*, label=-, labelsep=0.3em, itemsep=0pt, topsep=0pt, partopsep=0pt, parsep=0pt]
\item of workers were forced labourers according to a
\item dies Officers found two people working at the site who they believed were the victims of modern slavery A
\item trafficking These victims of organised crime have been found on remote cannabis farms in
\item network Bul garian nationals from disadvantaged regions were recruited by the traffickers to
\item 7 2 hours deep in the mines shafts ,
\end{itemize} & \begin{itemize}[leftmargin=*, label=-, labelsep=0.3em, itemsep=0pt, topsep=0pt, partopsep=0pt, parsep=0pt]
\item citizens
\item forced to
\item but not
\item cotton
\item Modern Day Slavery
\item UK
\item https :/
\item forced
\item An appeal has gone
\item rural
\item Wales
\item gang masters
\end{itemize} \\
\bottomrule
\end{tabular}

\end{table*}

\begin{table*}[h]
\footnotesize
\caption{Qualitative examples of extracted tokens across models (rows) and  human-annotated rationales, model-generated rationales, and post-hoc attribution-based (LRP) rationale for the RaFoLa dataset (here: article \#8). Tokens are selected from the top 5\% most faithful tokens in randomly selected samples.}

\label{tab:rafola_samples_art8}
\centering
\setlength{\tabcolsep}{1.8pt} 

\begin{tabular}{l p{5.5cm} p{5.5cm} p{3.6cm}}
\toprule
model & human & model & post-hoc \\
\midrule
\midrule
Llama3 & \begin{itemize}[leftmargin=*, label=-, labelsep=0.3em, itemsep=0pt, topsep=0pt, partopsep=0pt, parsep=0pt]
\item the blood ran out of the g ashes Fast
\item Many of the children also undergo physical and sexual abuse from the traff ickers as well as sometimes being forced into drug addiction
\item He threatened to kill my whole
\item investigators observed children working in the fields interviewed women who were paid
\item supervisors . The research show that in Jordan woman migrants routinely face sexual harassment and physical assaults by male supervisors . All
\end{itemize} & \begin{itemize}[leftmargin=*, label=-, labelsep=0.3em, itemsep=0pt, topsep=0pt, partopsep=0pt, parsep=0pt]
\item Many of the children also undergo physical and sexual abuse from the traff ickers as well as sometimes being forced into drug addiction
\item fields and sexually assaulted by plantation fore
\item reported being raped while working . The
\item Naz ma Ak ter told New Age that in
\item will quite often be exposed to physical
\end{itemize} & \begin{itemize}[leftmargin=*, label=-, labelsep=0.3em, itemsep=0pt, topsep=0pt, partopsep=0pt, parsep=0pt]
\item humiliation
\item later
\item them
\item from
\item sexual violence
\item Speaking
\item workers
\item Meanwhile
\item 16
\item girl describes how
\item raped her
\item .com
\end{itemize} \\
\midrule
Mistral & \begin{itemize}[leftmargin=*, label=-, labelsep=0.3em, itemsep=0pt, topsep=0pt, partopsep=0pt, parsep=0pt]
\item Many of the children also under go physical and sexual abuse from the traff ick ers as well as sometimes being forced into drug addiction
\item girl describes how her boss rap ed her amid the tall
\item AP investig ators observed children working in the fields interviewed women who were paid nothing and women and
\item The research show that in Jordan woman migr ants rout inely face sexual harass ment and physical assault s by male super vis ors 0x0A All
\item reported harsh pun ish ments for min ers not comp lying with the rules imposed by the criminal
\end{itemize} & \begin{itemize}[leftmargin=*, label=-, labelsep=0.3em, itemsep=0pt, topsep=0pt, partopsep=0pt, parsep=0pt]
\item Many of the children also under go physical and sexual abuse from the traff ick ers
\item girl who described being rap ed by her boss in
\item AP investig ators observed children working in the fields
\item The research show that in Jordan woman migr ants rout inely face sexual harass ment
\item In addition to severe beat ings other san ctions have included being shot in the hands or having a hand cut off as well as kill ings
\end{itemize} & \begin{itemize}[leftmargin=*, label=-, labelsep=0.3em, itemsep=0pt, topsep=0pt, partopsep=0pt, parsep=0pt]
\item Beg
\item Iran
\item earn
\item 0
\item abuse
\item year
\item describes how
\item boss rap ed her
\item 0x0A
\item worked
\item child labour
\item slavery
\end{itemize} \\
\midrule
Qwen3 & \begin{itemize}[leftmargin=*, label=-, labelsep=0.3em, itemsep=0pt, topsep=0pt, partopsep=0pt, parsep=0pt]
\item Many of the children also undergo physical and sexual abuse from the traff ickers as well as sometimes being forced into drug addiction
\item work . A quarter of those surveyed had experienced verbal
\item Come sleep with me I will give you a baby Now
\item They to il for hours a day with little to no food and face abuse at the hands of their masters
\item girl describes how her boss raped
\end{itemize} & \begin{itemize}[leftmargin=*, label=-, labelsep=0.3em, itemsep=0pt, topsep=0pt, partopsep=0pt, parsep=0pt]
\item Many of the children also undergo physical and sexual abuse from the traff ickers as well as sometimes being forced into drug addiction
\item 9 8 0 . Four in ten reported
\item 1 2 being taken into the fields and sexually assaulted by plantation fore men . While
\item can hear the sound of tools hitting stone of men cough ing
\item 1 2 being taken into the fields and sexually assaulted by plantation fore men . While
\end{itemize} & \begin{itemize}[leftmargin=*, label=-, labelsep=0.3em, itemsep=0pt, topsep=0pt, partopsep=0pt, parsep=0pt]
\item Iranian
\item Speaking
\item Iran
\item .
\item migrant
\item Risk
\item rapport
\item ".
\item the
\item palm oil plantation that
\item some
\item .
\end{itemize} \\
\midrule
Gemma3 & \begin{itemize}[leftmargin=*, label=-, labelsep=0.3em, itemsep=0pt, topsep=0pt, partopsep=0pt, parsep=0pt]
\item “ he would have them tied up to a
\item Many of the children also undergo physical and sexual abuse from the traffickers as well as sometimes being forced into drug addiction
\item According to the Typ ology study traffickers more frequently use physical violence in outdoor solicitation than in other types of sex trafficking Residential
\item officers of some garment factories h url abusive words at female workers and even go for sexual harassment If
\item work A quarter of those surveyed had experienced verbal
\end{itemize} & \begin{itemize}[leftmargin=*, label=-, labelsep=0.3em, itemsep=0pt, topsep=0pt, partopsep=0pt, parsep=0pt]
\item Many of the children also undergo physical and sexual abuse from the traffickers as well as sometimes being forced into drug addiction
\item It is a broad term used in a commercial sex trade referring to commercial sex acts This acts primarily occur at a temporary indoor location
\item officers of some garment factories h url abusive words at female workers and even go for sexual
\item € 9 8 0 Four in ten reported feeling unsafe at work
\item girl describes how her boss raped her amid the tall trees on an Indonesian palm oil plantation
\end{itemize} & \begin{itemize}[leftmargin=*, label=-, labelsep=0.3em, itemsep=0pt, topsep=0pt, partopsep=0pt, parsep=0pt]
\item beat
\item blood
\item Copyright
\item forced
\item beggars
\item . Meanwhile
\item In
\item ’
\item “
\item –
\item or
\item does not recognise
\end{itemize} \\
\bottomrule
\end{tabular}

\end{table*}

\end{document}